\documentclass{article}
\usepackage[utf8]{inputenc}
\usepackage[T1]{fontenc}
\usepackage{url}
\usepackage{booktabs}
\usepackage{placeins}
\usepackage{amsfonts}
\usepackage{nicefrac}
\usepackage{microtype}
\usepackage{xcolor}
\usepackage{graphicx}
\usepackage{flafter}
\usepackage{amsmath}
\usepackage{amssymb}
\usepackage{mathtools}
\usepackage{amsthm}
\usepackage{algorithm}
\usepackage{algorithmic}
\usepackage{tikz}
\usepackage{pgfplots}
\pgfplotsset{compat=1.17}
\usetikzlibrary{arrows.meta,decorations.pathreplacing}
\usepackage{hyperref}
\let\appendixaddcontentsline\addcontentsline
\usepackage[preprint]{icml2027/icml2027}
\usepackage[capitalize,noabbrev]{cleveref}

\usepackage{aliascnt}
\newcommand{\sharedtheorem}[2]{%
  \newaliascnt{#1}{theorem}\newtheorem{#1}[#1]{#2}\aliascntresetthe{#1}}
\theoremstyle{plain}

\sharedtheorem{proposition}{Proposition}
\sharedtheorem{lemma}{Lemma}
\sharedtheorem{corollary}{Corollary}
\theoremstyle{definition}
\sharedtheorem{definition}{Definition}
\sharedtheorem{assumption}{Assumption}
\theoremstyle{remark}
\sharedtheorem{remark}{Remark}
\crefname{assumption}{Assumption}{Assumptions}
\Crefname{assumption}{Assumption}{Assumptions}
\usepackage{thm-restate}

\newcommand{\PP}{\mathbb P}
\newcommand{\QQ}{\mathbb Q}
\newcommand{\EE}{\mathbb E}
\newcommand{\PX}{\PP_X}

\newcommand{\1}{\mathbf 1}
\newcommand{\KL}{\operatorname{KL}}
\newcommand{\calK}{\mathcal K}

\icmltitlerunning{Flow Partition Trees}

\begin{document}

\twocolumn[
  \icmltitle{Flow Partition Trees: Boosting Conditional Densities with Joint-Space Partitions}

  \begin{icmlauthorlist}
    \icmlauthor{Felipe Angelim}{fel}
  \end{icmlauthorlist}
  \icmlaffiliation{fel}{Independent Researcher}
  \icmlcorrespondingauthor{Felipe Angelim}{felipeangelim@pm.me}
  \icmlkeywords{Conditional Density Estimation, Probabilistic Regression,
Tree Ensembles, Boosting, Normalizing Flows}

  \vskip 0.3in
]

\printAffiliationsAndNotice{}

\begin{abstract}
A correctly specified conditional cumulative distribution function (CDF)
transforms a continuous response into a uniform variable at each covariate
value. We introduce Flow Partition Trees (FPT), a boosting method that
estimates conditional densities by correcting departures from this uniformity.
Each round fits a Partition Tree that splits on both the covariates and the transformed
response, then composes the tree's conditional CDF with the current model's CDF.
The resulting model gives direct access to conditional densities and
quantiles. We show how the tree's splitting criterion yields a lower bound
on the reduction in population log loss at each round.
We further extend FPT to fit the probability assigned to intervals,
particularly in the case of finite-precision continuous outcomes.
On real and synthetic regression tasks, tuned FPT is competitive with
leading tree-based methods for probabilistic prediction.
FPT's default configuration achieves the best average negative log-likelihood 
rank on the real-data benchmarks.
\end{abstract}

\section{Introduction}
\label{sec:intro}

Decision-making often requires an outcome's probability distribution,
not just its expected value. Supervised probabilistic regression
has applications in astronomy \citep{almosallam2016gpz},
weather forecasting \citep{gneiting2005calibrated, rasp2018neural},
predictive maintenance \citep{richardson2017gaussian}
and credit \citep{somers2007quantile, wattanawongwan2023modelling, matuszyk2010modelling}.

Tree algorithms are widely used for regression because of their training
speed and predictive performance. However, probabilistic tree methods
provide different forms of distributional information. Some estimate the
parameters of a specified family, such as Gaussian distributions;
others provide predictive samples or estimate selected quantiles.
These approaches can be restrictive when the goal is to evaluate densities,
probabilities and quantiles from a single flexible conditional distribution.

Recent tree methods estimate flexible conditional distributions directly.
CDTree \citep{yang2024conditional} partitions the covariate space and assigns
a histogram to each leaf. Partition Tree \citep{angelim2026partition}
partitions both the covariate and outcome space, estimating a density from
joint probability masses and response widths. Conditional Tree Flow
\citep{wang2024generative} composes tree-based transforms whose
covariate-dependent branch probabilities are learned by binary classifiers.
These constructions raise a question: can joint-space partitioning serve
as an effective residual learner for conditional CDF boosting, with a
partition objective that explains the improvement from each round?

We introduce Flow Partition Trees (FPT), using Partition Trees as base
learners in the probabilistic boosting framework of
\citet{gressmann2018probabilistic}. A correctly specified conditional CDF
transforms a continuous response into a uniform variable at each covariate
value. FPT sequentially fits Partition Trees to the transformed responses
and composes their conditional CDFs. Each tree allocates splits across
covariates and transformed responses to model the remaining departure
from conditional uniformity. The ensemble provides conditional densities,
CDFs and quantiles through the fitted transports.

The same construction accommodates responses recorded to finite precision.
Pointwise likelihood can reward repeated concentration at recorded values,
even with a fixed minimum leaf width in transported coordinates. We instead
fit the probability assigned to each recorded interval, using posterior
fractional counts to train the residual trees. We connect both the
continuous-response and interval-likelihood objectives to the divergence
captured by a partition.

Across ten real-data datasets, FPT achieves the best average CRPS rank,
with performance not statistically distinguished from Treeffuser or
CatBoost-R by the adjusted comparisons. Its default configuration achieves
the best average NLL rank. FPT also has lower recorded training and
prediction times than Treeffuser. Synthetic experiments and ablations
characterize its predictive behavior and the contributions of averaging,
randomized split candidates and target smoothing.
Our contributions are three-fold:
\begin{enumerate}
  \item We develop a conditional CDF boosting method with joint-space
  Partition Tree residual learners, providing directly evaluable densities,
  CDFs and quantiles, and establish its competitive predictive performance
  through real-data and synthetic benchmarks.
  \item We characterize the raw Partition Tree objective through a Bregman
  decomposition and show that its population partition score lower bounds
  per-round log-loss improvement after conditional normalization and shrinkage.
  \item We extend the learner to interval-valued observations, yielding a generalized
  expectation--maximization update with a bin-likelihood improvement bound.
\end{enumerate}

Source code is available at
\url{https://github.com/felipeangelimvieira/partition_tree}.

\section{Related work}
\label{sec:related-work}

\paragraph{Boosting and sample transport}
\citet{gressmann2018probabilistic} introduce probabilistic boosting based
on transforming responses through fitted conditional CDFs.
\citet{awaya2024unsupervised} develop unconditional tree boosting through
CDF transforms and connect residualization to sequential KL reduction.
FPT adopts this composition principle; its contributions concern the
joint-space residual learner, the connection between its partition score
and conditional log-loss improvement, and interval-likelihood fitting.

\paragraph{Conditional tree flows}
\citet{wang2024generative} construct conditional density models by composing
tree-based piecewise-linear transforms, learning covariate-dependent branch
probabilities through binary classification at outcome-tree splits.
FPT shares the transport framework and direct density evaluation, but fits
each residual layer by partitioning the joint space of covariates and
transformed responses. Leaf values are determined by joint and reference
masses. Our analysis links this partition objective to population improvement
after conditional normalization and shrinkage, and extends the construction
to quantized outcome measurements.

\paragraph{Partition-based conditional densities}
CDTree \citep{yang2024conditional} fits a covariate partition with a response
histogram in each leaf using a minimum description length objective.
Partition Tree \citep{angelim2026partition} supplies the joint-space base
learner used here. FPT builds on that learner through sequential CDF
composition; the boosting construction, improvement analysis and
interval-likelihood extension are the focus of this work.

\paragraph{Tree boosting for probabilistic regression}
CatBoost \citep{prokhorenkova2018catboost} supports Gaussian uncertainty
estimates \citep{malinin2020uncertainty}; NGBoost \citep{duan2020ngboost}
fits the parameters of a chosen conditional distribution using natural
gradients for a proper scoring rule. LinCDE \citep{gao2022lincde} combines
gradient boosting with Lindsey's method to fit a flexible density family
that can represent changes in modality and shape. It estimates density
parameters through boosted trees, whereas FPT composes the CDFs of
partition-based densities. Treeffuser \citep{beltranvelez2024treeffuser}
uses boosted trees to learn the score of a diffusion process and generates
predictive samples by numerically solving a reverse-time stochastic
differential equation. FPT instead evaluates its density and CDF directly
and computes quantiles by inverting the fitted piecewise-linear CDF.

\section{Methodology}
\label{sec:methodology}
FPT uses Partition Trees \citep{angelim2026partition} as base learners
in probabilistic regression boosting \citep{gressmann2018probabilistic}.

\subsection{Flow Partition Tree}
\label{sec:fpt}
\paragraph{Setup and notation}Given observations $\mathcal{D} = \{(x_i, y_i)\}_{i=1}^n$ drawn i.i.d. from an 
unknown joint distribution $\mathbb{P}_{XY}$ on $\mathcal{X} \times \mathcal{Y}$,
our goal is to estimate the conditional density $f(y \mid x)$ of $Y$ given $X=x$.
We assume that $\mathcal{Y} \subseteq \mathbb{R}$ and the covariates $X \in \mathcal{X}$
are numeric or categorical. We denote the true conditional CDF by $F(y \mid x)$.
 
\subsubsection{One Partition Tree}
A Partition Tree divides $\mathcal X\times\mathcal Y$ into rectangular
leaves $A=A_X\times A_Y$, providing a leaf-constant approximation
to the Radon-Nikodym derivative $d\PP_{XY}/d(\PX\otimes\mu)$
where $\mu$ denotes Lebesgue measure on $\mathbb R$.
On leaf $A$, the raw tree $f_\pi$ takes value $c_A$ at the population level
and $\hat c_A$ when fitted to data, where
\begin{equation}
  \label{eq:leaf-constant}
 c_A=\frac{\PP_{XY}(A)}{\PX(A_X)\mu(A_Y)},\qquad
 \hat{c}_A=\frac{\widehat\PP_{XY}(A)}{\widehat\PP_X(A_X)\mu(A_Y)}.
\end{equation}
The empirical numerator is the fraction of
observations in the leaf; the denominator is its covariate mass times
its response width. For a query $x$, the fitted conditional density and CDF are
\begin{align}
\label{eq:pt-pdfcdf}
 \bar f_\pi(y\mid x)
 &=\frac{\sum_{A\in\pi}\1_{(x,y)\in A}\hat{c}_A}{\varsigma_\pi(x)},\\
 \bar F_\pi(y\mid x)
 &=\frac{\sum_{A:x\in A_X}\mu(A_Y\cap(-\infty,y])\hat{c}_A}{\varsigma_\pi(x)},
\end{align}
where $\varsigma_\pi(x)=\sum_{A:x\in A_X}\mu(A_Y)\hat{c}_A>0$ normalizes
the density. A query traverses the tree using the covariate splits and
collects the response intervals from the remaining leaves.

\subsubsection{One flow path}
\label{sec:flow-path}
Flow Partition Tree (FPT) builds a sequence of Partition Trees $\{\bar f_{\pi_t}\}_{t=1}^T$
by sequentially transporting the targets through the CDF of each round,
following the boosting algorithm of \citet{gressmann2018probabilistic}.
Let $u_i^{(t)}$ denote the the transported target $y_i$ after $t$ rounds.
At initialization round 0, we choose a CDF $F_0: \mathcal{Y} \to [0,1]$ with density $f_0$,
independent of $x$, and set $u_i^{(0)} = F_0(y_i)$.
At each boosting round $t=1,\dots,T$, we fit a Partition Tree
$f_{\pi_t}$ to the transported dataset
$\mathcal{D}^{(t-1)} = \{(x_i, u_i^{(t-1)})\}_{i=1}^n$.
Its normalizer in transported coordinates is
$\varsigma_t(x)=\int_0^1 f_{\pi_t}(x,v)\,dv$. Shrinking the normalized
density $\bar f_{\pi_t}(u\mid x)=f_{\pi_t}(x,u)/\varsigma_t(x)$ towards
the uniform density on $[0,1]$ with learning rate $\alpha\in[0,1]$ gives
\[
  \tilde{f}_{\pi_t}(u \mid x)
  = (1-\alpha) + \alpha\,\bar f_{\pi_t}(u \mid x).
\]
Let $\tilde{F}_{\pi_t}$ denote its CDF. We then update the targets as
$u_i^{(t)} = \tilde{F}_{\pi_t}(u_i^{(t-1)} \mid x_i)$.

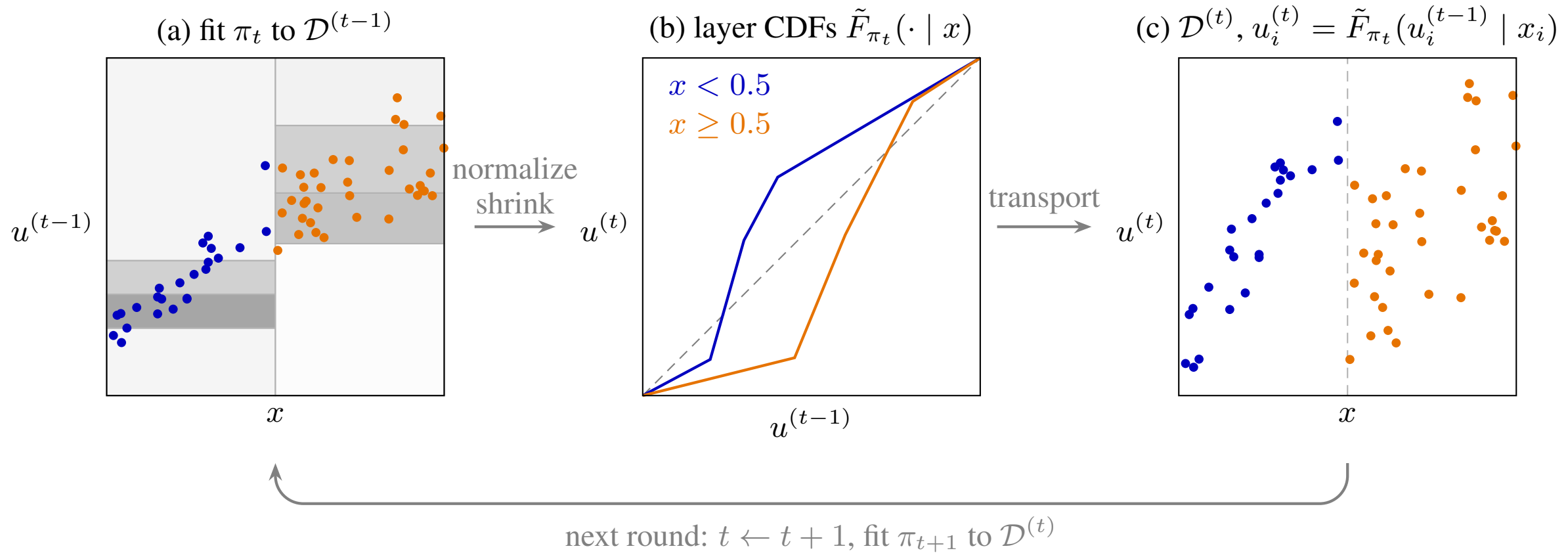
\begin{figure*}[t]
\centering
\begin{tikzpicture}[font=\small,x=3.4cm,y=3.4cm,
  pt/.style={circle,inner sep=0pt,minimum size=2.6pt},
  flowarrow/.style={-{Stealth[length=5pt]},thick,gray}]
\colorlet{c0}{blue!75!black}\colorlet{c1}{orange!90!black}
\foreach \xa/\xb/\ua/\ub/\sh/\v in {0/.5/0/.2/4/0.4, 0/.5/.2/.3/35/4.2, 0/.5/.3/.4/18/2.1, 0/.5/.4/1/4/0.5, .5/1/0/.45/1/0.1, .5/1/.45/.6/23/2.8, .5/1/.6/.8/18/2.2, .5/1/.8/1/5/0.6} {
  \filldraw[fill=black!\sh,draw=gray!60] (\xa,\ua) rectangle (\xb,\ub);
}
\foreach \x/\u in {0.238/0.286,0.259/0.359,0.47/0.681,0.151/0.292,0.301/0.472,0.473/0.486,0.395/0.438,0.217/0.334,0.301/0.395,0.163/0.286,0.285/0.452,0.089/0.261,0.151/0.242,0.294/0.374,0.044/0.157,0.331/0.407,0.31/0.436,0.031/0.238,0.197/0.256,0.156/0.318,0.042/0.243,0.238/0.288,0.02/0.178,0.06/0.2} \node[pt,fill=c0] at (\x,\u) {};
\foreach \x/\u in {0.604/0.512,0.626/0.556,0.837/0.667,0.836/0.523,0.635/0.616,0.741/0.528,0.671/0.699,0.591/0.576,0.719/0.696,0.921/0.593,0.936/0.607,0.879/0.728,0.965/0.592,0.507/0.43,0.585/0.569,0.584/0.617,0.929/0.622,0.856/0.818,0.861/0.882,0.569/0.477,0.714/0.632,0.574/0.654,0.99/0.828,0.615/0.659,0.881/0.803,0.999/0.733,0.959/0.659,0.897/0.612,0.52/0.541,0.644/0.468,0.62/0.483,0.941/0.606,0.72/0.592,0.521/0.674,0.548/0.578,0.58/0.525} \node[pt,fill=c1] at (\x,\u) {};
\draw (0,0) rectangle (1,1);
\node[below] at (0.5,0) {$x$};
\node[left] at (0,0.5) {$u^{(t-1)}$};
\node[above,align=center] at (0.5,1) {(a) fit $\pi_t$ to $\mathcal D^{(t-1)}$};
\draw[flowarrow] (1.09,0.5) -- node[above,align=center,font=\footnotesize] {normalize\\shrink} (1.33,0.5);
\begin{scope}[xshift=5.4cm]
\draw[densely dashed,gray] (0,0) -- (1,1);
\draw[thick,c0] (0,0) -- (0.2,0.107) -- (0.3,0.46) -- (0.4,0.647) -- (1,1);
\draw[thick,c1] (0,0) -- (0.45,0.112) -- (0.6,0.476) -- (0.8,0.871) -- (1,1);
\draw (0,0) rectangle (1,1);
\node[below] at (0.5,0) {$u^{(t-1)}$};
\node[left] at (0,0.5) {$u^{(t)}$};
\node[above,align=center] at (0.5,1) {(b) layer CDFs $\tilde F_{\pi_t}(\cdot\mid x)$};
\node[c0,anchor=west,font=\footnotesize] at (0.04,0.92) {$x<0.5$};
\node[c1,anchor=west,font=\footnotesize] at (0.04,0.8) {$x\geq0.5$};
\draw[flowarrow] (1.05,0.5) -- node[above,font=\footnotesize] {transport} (1.33,0.5);
\end{scope}
\begin{scope}[xshift=10.8cm]
\draw[densely dashed,gray!60] (0.5,0) -- (0.5,1);
\foreach \x/\u in {0.238/0.409,0.259/0.57,0.47/0.812,0.151/0.431,0.301/0.689,0.473/0.697,0.395/0.669,0.217/0.524,0.301/0.638,0.163/0.411,0.285/0.677,0.089/0.321,0.151/0.255,0.294/0.599,0.044/0.084,0.331/0.651,0.31/0.668,0.031/0.24,0.197/0.304,0.156/0.493,0.042/0.258,0.238/0.418,0.02/0.095,0.06/0.108} \node[pt,fill=c0] at (\x,\u) {};
\foreach \x/\u in {0.604/0.261,0.626/0.369,0.837/0.608,0.836/0.29,0.635/0.506,0.741/0.3,0.671/0.672,0.591/0.418,0.719/0.665,0.921/0.46,0.936/0.489,0.879/0.728,0.965/0.457,0.507/0.107,0.585/0.4,0.584/0.509,0.929/0.518,0.856/0.883,0.861/0.924,0.569/0.177,0.714/0.54,0.574/0.583,0.99/0.889,0.615/0.592,0.881/0.873,0.999/0.739,0.959/0.592,0.897/0.499,0.52/0.333,0.644/0.156,0.62/0.193,0.941/0.487,0.72/0.456,0.521/0.623,0.548/0.422,0.58/0.293} \node[pt,fill=c1] at (\x,\u) {};
\draw (0,0) rectangle (1,1);
\node[below] at (0.5,0) {$x$};
\node[left] at (0,0.5) {$u^{(t)}$};
\node[above,align=center] at (0.5,1) {(c) $\mathcal D^{(t)}$, $u_i^{(t)}=\tilde F_{\pi_t}(u_i^{(t-1)}\mid x_i)$};
\end{scope}
\draw[flowarrow,rounded corners=8pt] (10.8cm+0.5*3.4cm,-0.2) -- ++(0,-0.12)
  -- node[below,font=\footnotesize] {next round: $t\gets t+1$, fit $\pi_{t+1}$ to $\mathcal D^{(t)}$} (0.5,-0.32) -- (0.5,-0.2);
\end{tikzpicture}
\caption{One round of Flow Partition Tree on simulated data, colored
by covariate cell. (a) The tree splits $x$ at $0.5$ and then splits
the response differently in each cell; darker leaves have larger
values $\hat c_A$. (b) Normalizing and shrinking with
$\alpha=0.8$ gives one piecewise-linear CDF per covariate cell; the
dashed diagonal is the identity ($\alpha=0$). (c) Each response moves
through the CDF of its cell. Within each cell the transported responses
are closer to uniform, but the trend in $x$ inside the left cell
remains for the next tree to fit. The flow composes these maps, so
$\Psi_T$ and $\psi_T$ follow from \eqref{eq:fpt}.}
\label{fig:fpt-round}
\end{figure*}

We denote the composition of CDFs by $\Psi_t$, with $\Psi_0=F_0$:
\[
 \Psi_t(y\mid x)=\tilde F_{\pi_t}\bigl(\Psi_{t-1}(y\mid x)\mid x\bigr).
\]
The density $\psi_T$ after $T$ rounds follows from the chain rule:
\begin{equation}
\label{eq:fpt}
 \psi_T(y\mid x)=f_0(y)\prod_{t=1}^T
 \tilde f_{\pi_t}\bigl(\Psi_{t-1}(y\mid x)\mid x\bigr).
\end{equation}
\Cref{alg:prob-boosting} summarizes the procedure and
\cref{fig:fpt-round} illustrates one round.

In practice, for initialization, we use the uniform density
$f_0(y) = \1_{y \in [a_0, b_0]}/(b_0-a_0)$, where $a_0 < b_0$ are chosen
so that the interval contains all responses in $\mathcal{D}$.
If $\mathcal{Y}$ is bounded, we use its lower and upper bounds.

\paragraph{Residual density.}
Each tree estimates a correction to the current model. 
Let $\PP_{t-1}$ be the law of
$(X,U^{(t-1)})$, where $U^{(t-1)}=\Psi_{t-1}(Y\mid X)$.
Let $\QQ_u=\PX\otimes\mathrm{Unif}[0,1]$ be the reference law in transported
coordinates. At round $t$, the tree estimates $r_{t-1}=d\PP_{t-1}/d\QQ_u$.
By change of variables,
\begin{equation}
\label{eq:residual-ratio}
r_{t-1}\bigl(x, \Psi_{t-1}(y\mid x)\bigr) =\frac{f(y\mid x)}{\psi_{t-1}(y\mid x)}.
\end{equation}
A value above one indicates underestimation by the current model; a
value below one indicates overestimation. By \eqref{eq:fpt}, an exact
fit to this ratio with $\alpha=1$ recovers $f$ in one round. 
On a finite partition $\pi_t$, the population tree $f_{\pi_t}$ approximates
$r_{t-1}$ by its leaf averages.

\begin{algorithm}[t]
\caption{Greedy probabilistic residual boosting \citep{gressmann2018probabilistic}.}
\label{alg:prob-boosting}
\begin{algorithmic}[1]
\REQUIRE Data $\mathcal{D} = \{(x_i, y_i)\}_{i=1}^n \subset \mathcal{X} \times \mathcal{Y}$;
Partition Tree learners $f_{\pi_1}, \dots, f_{\pi_T}$ trainable on
$\mathcal{X} \times [0,1]$; initial CDF $F_0$ with density $f_0$;
learning rate $\alpha \in [0,1]$.
\ENSURE Conditional density estimator $\psi_T(\cdot \mid \cdot)$.
\STATE $u_i^{(0)} \gets F_0(y_i)$ for $i = 1, \dots, n$
\FOR{$t = 1$ \TO $T$}
  \STATE $\mathcal{D}^{(t-1)} \gets \{(x_i, u_i^{(t-1)})\}_{i=1}^n$
  \STATE Fit $f_{\pi_t}$ to $\mathcal{D}^{(t-1)}$ and normalize it at each $x$ to obtain $\bar f_{\pi_t}$
  \STATE $\tilde{f}_{\pi_t} \gets (1-\alpha) + \alpha\,\bar f_{\pi_t}$
  \STATE Let $\tilde{F}_{\pi_t}$ be the CDF of $\tilde{f}_{\pi_t}$
  \STATE $u_i^{(t)} \gets \tilde{F}_{\pi_t}(u_i^{(t-1)} \mid x_i)$ for $i = 1, \dots, n$
\ENDFOR
\RETURN $\psi_T(y \mid x)$ from \eqref{eq:fpt}.
\end{algorithmic}
\end{algorithm}

\subsection{Tree objective and per-round KL Improvement}
\label{sec:bregman}

Bregman divergence is commonly used in optimization problems to generalize the
notion of discrepancy between functions.
Given a differentiable, strictly convex generator
$\phi:(0,\infty)\to\mathbb R$, $\phi(1)=0$, define the Bregman divergence $d_\phi$ and its
expected value with respect to measure $\mathbb{Q}$
\[
 \begin{aligned}
 d_\phi(a,b)&=\phi(a)-\phi(b)-\phi'(b)(a-b),\\
 B_\phi(g,h)&=\int d_\phi(g,h)\,d\QQ.
 \end{aligned}
\]
At zero, $\phi$ is defined by its
limit from positive values, possibly infinite. For the logarithmic
generator $\phi(z)=z\log z$, this gives $0\log0=0$.
We demonstrate how
Partition Tree minimizes a Bregman divergence and in what sense
the leaf-constant in \Cref{eq:leaf-constant} is optimal at population
level.

Let $f$ be a density with respect to the probability measure $\QQ$.
For each leaf $A$ of a finite partition $\pi$, write
$q_A=\QQ(A)$, $p_A=\int_A f\,d\QQ$, and $c_A=p_A/q_A$.
Let $f_\pi=\sum_{A\in\pi}c_A\1_A$ be the tree of leaf averages and
$g_\pi=\sum_{A\in\pi}g_A\1_A$ a candidate tree. \Cref{thm:Bregman-decomposition} extends
the Theorem of \cite{banerjee2005clustering} for partition-based conditional density
population estimates.

\begin{restatable}[Bregman decomposition]{theorem}{bregmandecomposition}
\label{thm:Bregman-decomposition}
For a finite partition with $p_A,q_A>0$, positive leaf values $g_A$,
and $\int|\phi(f)|\,d\QQ<\infty$,
\begin{equation}
\label{eq:Bregman-decomposition}
 \begin{aligned}
 B_\phi(f,g_\pi)
 &=B_\phi(f,1)-B_\phi(f_\pi,1)\\
 &\quad+\sum_Aq_A d_\phi(c_A,g_A).
 \end{aligned}
\end{equation}
The unique optimal leaf values are $g_A=c_A$, giving
\begin{equation}
\label{eq:Bregman-decomposition-optimal-g_A}
 B_\phi(f,f_\pi)=B_\phi(f,1)-B_\phi(f_\pi,1).
\end{equation}
\end{restatable}
Proof can be found in \Cref{app:proofs}. The decomposition separates the loss of a raw tree into three terms.
The baseline $B_\phi(f,1)$ is the loss incurred by predicting the constant
density one. The partition captures $B_\phi(f_\pi,1)$ of this loss, and
$\sum_{A\in\pi}q_A d_\phi(c_A,g_A)$ is the additional loss incurred when
the fitted leaf values differ from the cell averages. Thus every generator
$\phi$ gives the same optimal values $g_A=c_A$ on a fixed partition.
The choice of $\phi$ determines how we score the partition:
\[
  B_\phi(f_\pi,1)=\sum_{A\in\pi}q_A\phi(c_A).
\]
For $\phi(z)=z\log z$, the increase in this score from splitting a leaf
is the population counterpart of the Partition Tree split gain
\citep{angelim2026partition}. At each node, the algorithm chooses an
admissible split with the largest empirical gain.

At round $t$, apply \cref{thm:Bregman-decomposition} with $f=r_{t-1}$
and reference law $\QQ_u$.
The population tree $f_{\pi_t}$ takes value $p_A^{(t-1)}/q_A$ on each
leaf $A\in\pi_t$, where $p_A^{(t-1)}=\PP_{t-1}(A)$. For $\phi(z)=z\log z$,
\begin{equation}
  \label{eq:divergence-step-t}
  D_t(\pi_t) := B_\phi(f_{\pi_t},1)
  = \sum_{A\in\pi_t} p_A^{(t-1)}\log\frac{p_A^{(t-1)}}{q_A}.
\end{equation}
The reduction in loss from fitting the raw tree is therefore exactly
the KL divergence between the leaf probabilities under $\PP_{t-1}$
and $\QQ_u$.

Let $\calK_t=\EE_{\PX}\KL(f(\cdot\mid X)\|\psi_t(\cdot\mid X))$ be the
excess expected log loss after $t$ rounds. By \eqref{eq:residual-ratio},
$\calK_t=B_\phi(r_t,1)$. \Cref{prop:flow-improvement} 
connects the Bregman objective of a single tree to the overall
improvement in KL-divergence of the ensemble.

\begin{restatable}[Per-round improvement]{proposition}{perroundimprovement}
\label{prop:flow-improvement}
Let $\phi(z)=z\log z$. Suppose that $f$ and $\psi_{t-1}$ are supported on the initial interval $I_0$,
$\psi_{t-1}>0$ there, and $\calK_{t-1}<\infty$. For a finite
partition $\pi_t$ with $p_A^{(t-1)},q_A>0$ on every leaf, the population
layer $\tilde f_{\pi_t}=(1-\alpha)+\alpha\bar f_{\pi_t}$ with $\alpha\in[0,1]$ satisfies
\begin{equation}
\label{eq:bregman-flow-progress}
 \begin{aligned}
 \calK_{t-1}-\calK_t
 &=\EE_{\PP_{t-1}}\log\tilde f_{\pi_t}(U^{(t-1)}\mid X)\\
 &\geq\alpha D_t(\pi_t) \geq0.
 \end{aligned}
\end{equation}
\end{restatable}

The proof is provided in
\cref{app:proofs}. \Cref{prop:flow-improvement}
shows that the same partition score used to choose splits gives a lower
bound on how much the next round reduces population log-loss.
If we assume a weak learning property on Partition Tree,
we obtain a geometric contraction of the KL-Divergence
during Flow Partition Tree boosting rounds, as in
\Cref{cor:geometric-decay}.

\begin{restatable}{corollary}{geometricdecay}
  \label{cor:geometric-decay}
  Assume the selected trees in Flow Partition Tree ensemble
  capture at least a fixed fraction $\gamma > 0$ of the
  remaining KL divergence at every round. Then:
  \begin{equation}
    \label{eq:kl-decay}
    D_t(\pi_t) \geq \gamma \calK_{t-1} \implies \calK_T \leq (1 - \alpha\gamma)^T\calK_0
  \end{equation}
\end{restatable}

\subsection{Target quantization}
\label{sec:quantized}

Rounding a continuous response identifies an interval containing its latent
value. Treating the recorded value as an exact point can reward a tree
for concentrating density at the same location in every round.
For example, suppose the covariates are constant and the recorded values
are $0$ and $1$ in proportions $4/5$ and $1/5$. Initialize FPT with the
uniform density on $[0,1]$, set $\alpha=1$, and require each target leaf
to have width at least $1/2$ in transported coordinates. Each round
splits at $1/2$ and assigns densities $8/5$ and $2/5$ to the two leaves.
The layer CDF fixes both endpoints, so after $T$ rounds the mean
pointwise log-likelihood is
\[
 T\{\tfrac45\log(8/5)+\tfrac15\log(2/5)\}\approx0.193T.
\]
Thus pointwise log-likelihood can increase without bound despite a fixed
minimum leaf width.

\paragraph{Bin log-likelihood.} Let $\delta>0$ be the measurement resolution and
$\bar y=\lfloor y\rceil_\delta\in\delta\mathbb Z$ the recorded response.
For $\mathcal D_\delta=\{(x_i,\bar y_i)\}_{i=1}^n$, define
$B(\bar y)=[\bar y-\delta/2,\bar y+\delta/2)$ and $B_i=B(\bar y_i)$.
Under a candidate density $\psi$, the probability of recording $\bar y$
at $x$ is the mass of its bin $B=B(\bar y)$,
\begin{equation}
\label{eq:quantized-density}
 m_\psi(B\mid x)=\int_B\psi(y\mid x)\,dy.
\end{equation}
The fitting objective is the bin log-likelihood
\begin{equation}
\label{eq:quantized-loglik}
 L_\delta(\psi)
 =\EE_{(X,\bar Y)}\log m_\psi(B(\bar Y)\mid X).
\end{equation}
Since each bin has probability at most one, $L_\delta(\psi)\leq 0$.
We fit this objective by generalized expectation--maximization (EM)
\citep{dempster1977maximum}, treating the response within each recorded bin as latent.

\begin{figure}[!htbp]
\centering
\begin{tikzpicture}[font=\small,
  declare function={pdf(\y)=12*\y*(1-\y)*(1-\y);
                    cdf(\y)=\y*\y*(6-8*\y+3*\y*\y);}]
\def\W{3.4}\def\H{3}\def\DB{-2.6}\def\DS{0.85}\def\XR{4}\def\US{1.1}
\def\yl{0.35}\def\yr{0.6}
\pgfmathsetmacro{\ul}{\H*cdf(\yl)}
\pgfmathsetmacro{\ur}{\H*cdf(\yr)}
\fill[blue!20] plot[domain=\yl:\yr,samples=20] ({\W*\x},{\DB+\DS*pdf(\x)})
  -- ({\W*\yr},\DB) -- ({\W*\yl},\DB) -- cycle;
\fill[blue!20] (\XR,\ul) rectangle (\XR+\US,\ur);
\foreach \y/\u in {\yl/\ul,\yr/\ur}
  \draw[densely dashed,gray] ({\W*\y},\DB) -- ({\W*\y},\u) -- (\XR+\US,\u);
\draw (0,\H) -- (0,0) -- (\W,0);
\node[left] at (0,0) {$0$};
\node[left] at (0,\H) {$1$};
\node[left] at (0,\H/2) {$u^{(t-1)}$};
\draw[thick] plot[domain=0:1,samples=60] ({\W*\x},{\H*cdf(\x)});
\node[above left] at ({\W*0.55},{\H*cdf(0.55)}) {$\Psi_{t-1}(y\mid x_i)$};
\draw[->] (0,\DB) -- (\W+0.25,\DB) node[right] {$y$};
\draw[thick] plot[domain=0:1,samples=60] ({\W*\x},{\DB+\DS*pdf(\x)});
\node[above right] at ({\W*0.8},{\DB+\DS*pdf(0.8)}) {$\psi_{t-1}(y\mid x_i)$};
\node[font=\scriptsize] at ({\W*(\yl+\yr)/2},\DB+0.45) {$m_i^{(t-1)}$};
\draw[decorate,decoration={brace,mirror,amplitude=3pt}]
  ({\W*\yl},\DB-0.08) -- ({\W*\yr},\DB-0.08) node[midway,below=3pt] {$B_i$};
\draw[gray] (\XR,0) rectangle (\XR+\US,\H);
\draw[->] (\XR,0) -- (\XR,\H+0.3) node[above] {$u^{(t-1)}$};
\node[below] at (\XR,0) {$0$};
\node[below] at (\XR+\US,0) {$1$};
\node[font=\scriptsize] at (\XR+\US/2,{(\ul+\ur)/2}) {$m_i^{(t-1)}$};
\draw[decorate,decoration={brace,amplitude=4pt}]
  (\XR+\US+0.08,\ur) -- (\XR+\US+0.08,\ul)
  node[midway,right=4pt,align=left] {$I_i^{(t-1)}$\\[-2pt]\scriptsize length $m_i^{(t-1)}$};
\end{tikzpicture}%
\caption{The shaded area under $\psi_{t-1}(\cdot\mid x_i)$ is the bin
probability $m_i^{(t-1)}$. The current CDF maps the bin endpoints to those
of $I_i^{(t-1)}$. In transported coordinates the same mass fills a rectangle
of unit density. Conditioning on the bin gives density $1/m_i^{(t-1)}$
throughout $I_i^{(t-1)}$.}

\label{fig:bin-transport}
\end{figure}
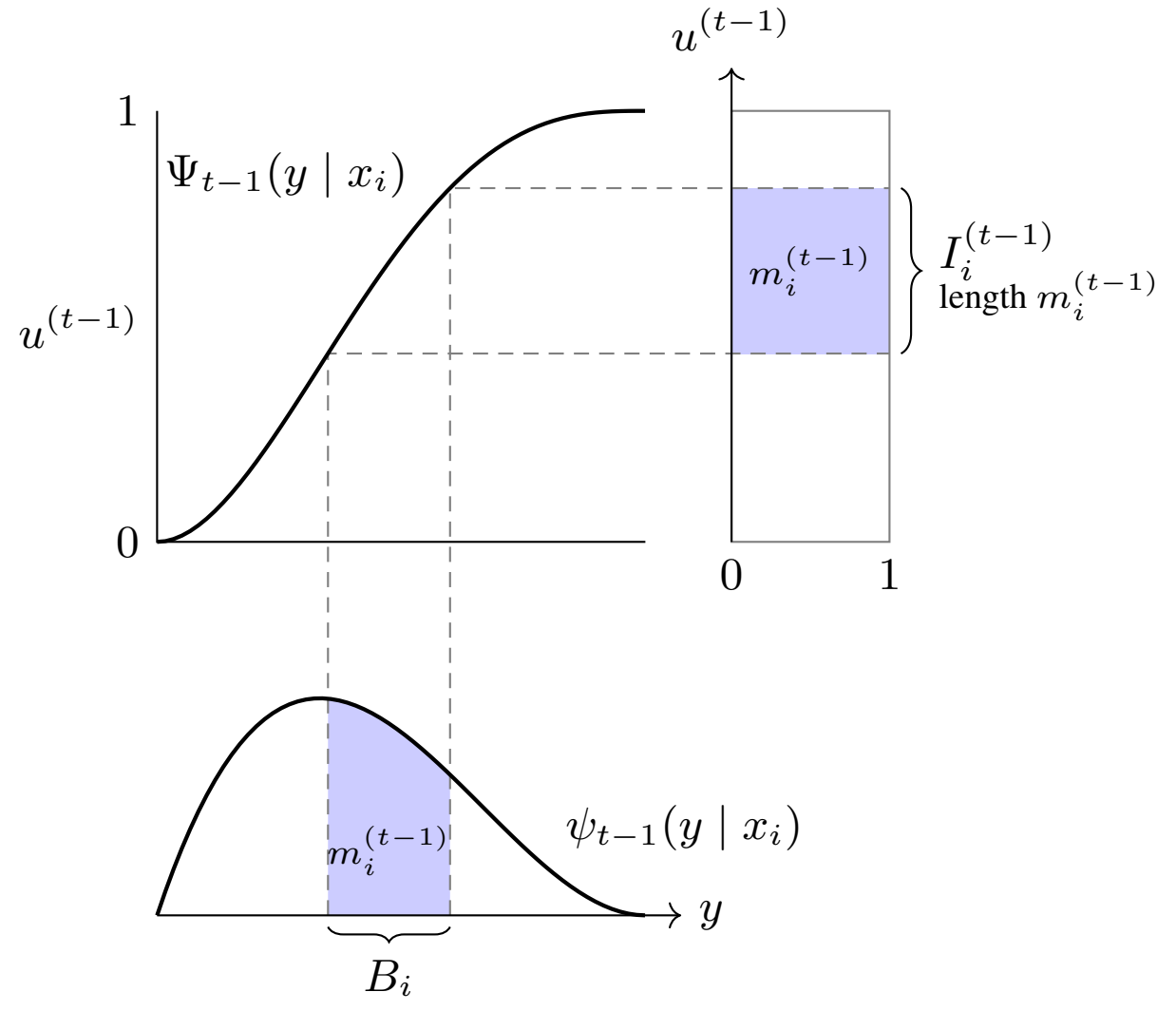

\paragraph{The E-step in transported coordinates.}
At the start of round $t$, condition the current model on observation
$i$'s bin $B_i=[\ell_i,r_i)$. If its mass is
$m_i^{(t-1)}=m_{\psi_{t-1}}(B_i\mid x_i)>0$, the posterior density of
the latent response on $B_i$ is $\psi_{t-1}(y\mid x_i)/m_i^{(t-1)}$.
The current CDF maps the bin to
\[
 I_i^{(t-1)}=\bigl[\Psi_{t-1}(\ell_i\mid x_i),\Psi_{t-1}(r_i\mid x_i)\bigr).
\]
The change of variables $u^{(t-1)}=\Psi_{t-1}(y\mid x_i)$ gives
\[
 m_i^{(t-1)}=\int_{B_i}\psi_{t-1}(y\mid x_i)\,dy
          =\mu(I_i^{(t-1)}).
\]
The same substitution on any subinterval $J\subseteq I_i^{(t-1)}$ gives it
mass $\mu(J)$. Conditional on the recorded bin, its probability is
therefore $\mu(J)/\mu(I_i^{(t-1)})$. Thus $U^{(t-1)}=\Psi_{t-1}(Y\mid x_i)$
is uniform on $I_i^{(t-1)}$ under the E-step posterior (\cref{fig:bin-transport}).

At the population level, let $p_B(x)$ be the probability of recording
bin $B$ given $X=x$, and let $m_B^{(t)}(x)=m_{\psi_t}(B\mid x)$ be its
model probability after $t$ rounds. Assume $m_B^{(t-1)}(x)>0$ whenever $p_B(x)>0$.
The E-step spreads mass $p_B(x)$ uniformly over the transported interval
$I_B^{(t-1)}(x)=\Psi_{t-1}(B\mid x)$ of length $m_B^{(t-1)}(x)$.

\paragraph{Divergence captured by the partition.}
Dividing the bin probability by its transported width gives the
density of the joint E-step law $\PP_{t-1,\delta}$ relative to $\QQ_u$,
\[
 r_{t-1,\delta}(x,u^{(t-1)})=\frac{p_B(x)}{m_B^{(t-1)}(x)},
 \qquad u^{(t-1)}\in I_B^{(t-1)}(x).
\]
This is the bin-level counterpart of the residual ratio in
\eqref{eq:residual-ratio}. A candidate leaf $A$ at round $t$ has E-step
mass $p_A^{(t-1)}=\PP_{t-1,\delta}(A)$ and reference mass $q_A=\QQ_u(A)$.
Under the hypotheses of \cref{thm:Bregman-decomposition}, the raw tree
$f_\pi$ with leaf values $p_A^{(t-1)}/q_A$ satisfies
\begin{equation}
\label{eq:bin-bregman-decomposition}
 B_\phi(r_{t-1,\delta},f_\pi)
 = B_\phi(r_{t-1,\delta},1)-\sum_{A\in\pi}q_A\phi(p_A^{(t-1)}/q_A).
\end{equation}
For $\phi(z)=z\log z$, the tree uses the unquantized objective $D_t(\pi)$
from \eqref{eq:divergence-step-t} with E-step leaf masses $p_A^{(t-1)}$.
The corollary in Appendix~\ref{app:proofs} shows that $\alpha D_t(\pi_t)$
lower bounds the bin log-likelihood gain at round $t$.

\paragraph{Fractional counts for tree splits.}
We can estimate $p_A^{(t-1)}$ directly from the recorded bins. Under the
uniform posterior on $I_i^{(t-1)}$, the probability that observation $i$
belongs to node $A=A_X\times A_U$ is
\begin{equation}
\label{eq:bin-membership}
 w_i^{(t-1)}(A)=\1_{x_i\in A_X}
 \frac{\mu(I_i^{(t-1)}\cap A_U)}{\mu(I_i^{(t-1)})}.
\end{equation}
A covariate split sends the entire parent weight to the child selected
by $x_i$. A split on the target coordinate that cuts through $I_i^{(t-1)}$ assigns weight to
both children in proportion to their overlaps. Their weights sum to $w_i^{(t-1)}(A)$.

Averaging these weights gives the empirical node mass. The reference
mass is the covariate mass times the response width. Thus
\[
 \widehat p_A^{(t-1)}=\frac1n\sum_{i=1}^n w_i^{(t-1)}(A),\qquad
 \widehat q_A=\widehat\PP_X(A_X)\mu(A_U).
\]
The tree fitted in round $t$ has raw leaf value
$\hat c_A^{(t)}=\widehat p_A^{(t-1)}/\widehat q_A$ as in \eqref{eq:leaf-constant}.
The empirical increase in captured divergence is the sum of the two
child terms in \eqref{eq:divergence-step-t} minus the parent's term
$\widehat p_A^{(t-1)}\log\hat c_A^{(t)}$. After normalizing and shrinking
the fitted tree, we apply its CDF to both endpoints of $I_i^{(t-1)}$
to obtain $I_i^{(t)}$ for the next round.

\subsection{Ensemble of Flow Paths}
\label{sec:bagging}
We consider a bagged configuration of FPT that averages the 
predictive distributions of multiple randomized flow paths.
The paths are trained in parallel using the following sources 
of randomization:
\begin{enumerate}
  \item Row subsampling: we give each FPT a subsample of
  the training rows
  \item Random candidate split points: considering that
  target splits dictate the shape of the data in future boosting
  steps, we add a randomization where the candidate split
  points for a split in the $Y$ coordinate are not defined
  by midpoints of consecutive observed values; instead,
  we randomly sample pairs and take their midpoint as candidate 
  split.
\end{enumerate}

\section{Experiments}
\label{sec:experiments}

We run the experiments to answer the following questions:
(a) how FPT compare with state-of-the-art probabilistic trees;
(b) how the ensemble and training choices contribute to performance;
and (c) how performance varies with dataset characteristics.
We denote by FPT-D the FPT with default hyperparameters.
FPT and FPT-D use nine flow paths. One searches target splits
exhaustively, and the other eight each use up to four random-pair
candidates per node.
We estimate the target bin width from the training responses in each fit
(\cref{app:bin-width}).

\subsection{Predictive performance on real data}
\label{sec:exp-real}

\paragraph{Datasets.}
We use the ten regression datasets in \cref{tab:univariate-full},
ranging from 768 to 45,730 observations.

\paragraph{Evaluation metrics.}
Our primary score is the continuous ranked probability score (CRPS).
For an observed response $y$, it integrates the squared difference
between the predicted CDF and $\1_{\{y\leq z\}}$ over thresholds $z$.
Lower scores are better. We also report negative
log-likelihood (NLL) for density fit, root mean squared error (RMSE)
for point prediction, and probabilistic calibration error (PCE), the
mean absolute gap between nominal and empirical coverage at the nine
predicted deciles. FPT evaluates its density directly, while Treeffuser's
NLL uses a kernel density estimate from predictive samples. We rank
methods by their mean score within each dataset and give each dataset
equal weight in the average rank.

\paragraph{Models and baselines.}
We compare FPT and FPT-D with Partition Tree and Partition Forest \citep{angelim2026partition},
Conditional Tree Flow \citep{wang2024generative}, LinCDE \citep{gao2022lincde},
NGBoost \citep{duan2020ngboost}, Treeffuser \citep{beltranvelez2024treeffuser},
and two Gaussian CatBoost configurations \citep{malinin2020uncertainty}.

\paragraph{Training and test procedure.}
On each dataset, all methods use the same shuffled five-fold
cross-validation splits.
For each tuned model, we select hyperparameters by CRPS on an 80/20
split of the training data within each fold. The search
ends after 100 trials or 20 minutes, with time checked between trials.
All FPT fits select the number of boosting rounds by validation
CRPS. In the tuned CatBoost comparisons, CatBoost uses
validation NLL for early stopping, while CatBoost-R uses validation
RMSE and a wider hyperparameter search. \Cref{app:hyperparameters} gives the
search spaces and model configurations.

\paragraph{Results.}
FPT has the best mean CRPS rank (2.1), followed by Treeffuser (2.3).
FPT has the lowest mean CRPS on
Abalone, Energy and Naval, while Treeffuser leads on four datasets,
including California Housing and Power Plant
(\cref{tab:univariate-full}). The Holm-adjusted
comparisons across datasets do not distinguish FPT from either
Treeffuser or CatBoost-R
(\cref{tab:univariate-main}) in terms of CRPS. FPT has lower mean CRPS
than Partition Tree on all ten datasets (\cref{fig:real-baselines-crps}).

\begin{table*}
\centering
\scriptsize
\setlength{\tabcolsep}{2pt}
\renewcommand{\arraystretch}{1.18}
\caption{CRPS: five-fold means, with sample SDs beneath in parentheses. A multiplier beside a dataset applies to both means and SDs in that row. Bold marks the lowest unrounded mean; underlined means are not significantly different from it (two-sided corrected resampled $t$-test over the five folds \citep{nadeau2003inference}, Holm-adjusted within the row \citep{holm1979simple}, $\alpha=0.05$). The final row averages within-dataset ranks across the ten datasets (lower is better); its marks follow \Cref{tab:univariate-main}. TF denotes Treeffuser, CB CatBoost, CB-R CatBoost-R, and NG NGBoost; remaining abbreviations follow \Cref{tab:univariate-main}.}
\label{tab:univariate-full}
\begin{tabular*}{\linewidth}{@{\extracolsep{\fill}}lrrrrrrrrrr@{}}
\toprule
Dataset & FPT & FPT-D & TF & CB & CB-R & NG & LinCDE & PT & PF & CTF \\
\midrule
Abalone & \shortstack[r]{\textbf{1.052}\\[-1pt]{\fontsize{6}{7}\selectfont(0.062)}} & \shortstack[r]{\underline{1.055}\\[-1pt]{\fontsize{6}{7}\selectfont(0.062)}} & \shortstack[r]{1.082\\[-1pt]{\fontsize{6}{7}\selectfont(0.064)}} & \shortstack[r]{\underline{1.112}\\[-1pt]{\fontsize{6}{7}\selectfont(0.039)}} & \shortstack[r]{\underline{1.107}\\[-1pt]{\fontsize{6}{7}\selectfont(0.068)}} & \shortstack[r]{1.098\\[-1pt]{\fontsize{6}{7}\selectfont(0.058)}} & \shortstack[r]{\underline{1.119}\\[-1pt]{\fontsize{6}{7}\selectfont(0.068)}} & \shortstack[r]{\underline{1.141}\\[-1pt]{\fontsize{6}{7}\selectfont(0.081)}} & \shortstack[r]{\underline{1.080}\\[-1pt]{\fontsize{6}{7}\selectfont(0.063)}} & \shortstack[r]{\underline{1.092}\\[-1pt]{\fontsize{6}{7}\selectfont(0.058)}} \\
Bike sharing & \shortstack[r]{\underline{16.40}\\[-1pt]{\fontsize{6}{7}\selectfont(0.62)}} & \shortstack[r]{\underline{16.69}\\[-1pt]{\fontsize{6}{7}\selectfont(0.61)}} & \shortstack[r]{\textbf{16.31}\\[-1pt]{\fontsize{6}{7}\selectfont(0.41)}} & \shortstack[r]{17.72\\[-1pt]{\fontsize{6}{7}\selectfont(0.66)}} & \shortstack[r]{\underline{17.14}\\[-1pt]{\fontsize{6}{7}\selectfont(0.70)}} & \shortstack[r]{17.55\\[-1pt]{\fontsize{6}{7}\selectfont(0.66)}} & \shortstack[r]{32.72\\[-1pt]{\fontsize{6}{7}\selectfont(0.81)}} & \shortstack[r]{30.66\\[-1pt]{\fontsize{6}{7}\selectfont(1.46)}} & \shortstack[r]{30.30\\[-1pt]{\fontsize{6}{7}\selectfont(1.18)}} & \shortstack[r]{68.18\\[-1pt]{\fontsize{6}{7}\selectfont(1.39)}} \\
California & \shortstack[r]{0.207\\[-1pt]{\fontsize{6}{7}\selectfont(0.005)}} & \shortstack[r]{0.208\\[-1pt]{\fontsize{6}{7}\selectfont(0.006)}} & \shortstack[r]{\textbf{0.194}\\[-1pt]{\fontsize{6}{7}\selectfont(0.004)}} & \shortstack[r]{0.236\\[-1pt]{\fontsize{6}{7}\selectfont(0.007)}} & \shortstack[r]{0.222\\[-1pt]{\fontsize{6}{7}\selectfont(0.005)}} & \shortstack[r]{0.224\\[-1pt]{\fontsize{6}{7}\selectfont(0.005)}} & \shortstack[r]{0.257\\[-1pt]{\fontsize{6}{7}\selectfont(0.006)}} & \shortstack[r]{0.273\\[-1pt]{\fontsize{6}{7}\selectfont(0.004)}} & \shortstack[r]{0.265\\[-1pt]{\fontsize{6}{7}\selectfont(0.012)}} & \shortstack[r]{0.327\\[-1pt]{\fontsize{6}{7}\selectfont(0.005)}} \\
Communities & \shortstack[r]{\underline{0.06436}\\[-1pt]{\fontsize{6}{7}\selectfont(0.00568)}} & \shortstack[r]{\underline{0.06448}\\[-1pt]{\fontsize{6}{7}\selectfont(0.00620)}} & \shortstack[r]{\underline{0.06420}\\[-1pt]{\fontsize{6}{7}\selectfont(0.00541)}} & \shortstack[r]{0.06912\\[-1pt]{\fontsize{6}{7}\selectfont(0.00680)}} & \shortstack[r]{\underline{0.06937}\\[-1pt]{\fontsize{6}{7}\selectfont(0.00743)}} & \shortstack[r]{\underline{0.06677}\\[-1pt]{\fontsize{6}{7}\selectfont(0.00580)}} & \shortstack[r]{\underline{0.06652}\\[-1pt]{\fontsize{6}{7}\selectfont(0.00528)}} & \shortstack[r]{\underline{0.07215}\\[-1pt]{\fontsize{6}{7}\selectfont(0.00794)}} & \shortstack[r]{\underline{0.06649}\\[-1pt]{\fontsize{6}{7}\selectfont(0.00564)}} & \shortstack[r]{\textbf{0.06419}\\[-1pt]{\fontsize{6}{7}\selectfont(0.00584)}} \\
Concrete & \shortstack[r]{\underline{2.237}\\[-1pt]{\fontsize{6}{7}\selectfont(0.200)}} & \shortstack[r]{\underline{2.296}\\[-1pt]{\fontsize{6}{7}\selectfont(0.211)}} & \shortstack[r]{\underline{2.284}\\[-1pt]{\fontsize{6}{7}\selectfont(0.106)}} & \shortstack[r]{\underline{2.763}\\[-1pt]{\fontsize{6}{7}\selectfont(0.226)}} & \shortstack[r]{\textbf{2.203}\\[-1pt]{\fontsize{6}{7}\selectfont(0.151)}} & \shortstack[r]{\underline{2.506}\\[-1pt]{\fontsize{6}{7}\selectfont(0.243)}} & \shortstack[r]{3.507\\[-1pt]{\fontsize{6}{7}\selectfont(0.242)}} & \shortstack[r]{4.804\\[-1pt]{\fontsize{6}{7}\selectfont(0.475)}} & \shortstack[r]{3.964\\[-1pt]{\fontsize{6}{7}\selectfont(0.474)}} & \shortstack[r]{4.910\\[-1pt]{\fontsize{6}{7}\selectfont(0.370)}} \\
Energy & \shortstack[r]{\textbf{0.198}\\[-1pt]{\fontsize{6}{7}\selectfont(0.020)}} & \shortstack[r]{\underline{0.233}\\[-1pt]{\fontsize{6}{7}\selectfont(0.013)}} & \shortstack[r]{\underline{0.224}\\[-1pt]{\fontsize{6}{7}\selectfont(0.048)}} & \shortstack[r]{\underline{0.258}\\[-1pt]{\fontsize{6}{7}\selectfont(0.052)}} & \shortstack[r]{\underline{0.212}\\[-1pt]{\fontsize{6}{7}\selectfont(0.021)}} & \shortstack[r]{\underline{0.207}\\[-1pt]{\fontsize{6}{7}\selectfont(0.017)}} & \shortstack[r]{0.995\\[-1pt]{\fontsize{6}{7}\selectfont(0.051)}} & \shortstack[r]{1.575\\[-1pt]{\fontsize{6}{7}\selectfont(0.089)}} & \shortstack[r]{1.611\\[-1pt]{\fontsize{6}{7}\selectfont(0.364)}} & \shortstack[r]{1.133\\[-1pt]{\fontsize{6}{7}\selectfont(0.103)}} \\
Kin8nm & \shortstack[r]{0.065\\[-1pt]{\fontsize{6}{7}\selectfont(0.002)}} & \shortstack[r]{0.068\\[-1pt]{\fontsize{6}{7}\selectfont(0.002)}} & \shortstack[r]{\underline{0.061}\\[-1pt]{\fontsize{6}{7}\selectfont(0.001)}} & \shortstack[r]{0.068\\[-1pt]{\fontsize{6}{7}\selectfont(0.001)}} & \shortstack[r]{\textbf{0.057}\\[-1pt]{\fontsize{6}{7}\selectfont(0.001)}} & \shortstack[r]{0.075\\[-1pt]{\fontsize{6}{7}\selectfont(0.002)}} & \shortstack[r]{0.085\\[-1pt]{\fontsize{6}{7}\selectfont(0.004)}} & \shortstack[r]{0.097\\[-1pt]{\fontsize{6}{7}\selectfont(0.002)}} & \shortstack[r]{0.088\\[-1pt]{\fontsize{6}{7}\selectfont(0.002)}} & \shortstack[r]{0.112\\[-1pt]{\fontsize{6}{7}\selectfont(0.002)}} \\
Naval {\tiny($10^{-3}$)} & \shortstack[r]{\textbf{0.372}\\[-1pt]{\fontsize{6}{7}\selectfont(0.010)}} & \shortstack[r]{\underline{0.379}\\[-1pt]{\fontsize{6}{7}\selectfont(0.011)}} & \shortstack[r]{\underline{0.401}\\[-1pt]{\fontsize{6}{7}\selectfont(0.022)}} & \shortstack[r]{0.508\\[-1pt]{\fontsize{6}{7}\selectfont(0.019)}} & \shortstack[r]{0.394\\[-1pt]{\fontsize{6}{7}\selectfont(0.014)}} & \shortstack[r]{0.467\\[-1pt]{\fontsize{6}{7}\selectfont(0.034)}} & \shortstack[r]{2.261\\[-1pt]{\fontsize{6}{7}\selectfont(0.183)}} & \shortstack[r]{2.737\\[-1pt]{\fontsize{6}{7}\selectfont(0.549)}} & \shortstack[r]{1.799\\[-1pt]{\fontsize{6}{7}\selectfont(0.286)}} & \shortstack[r]{2.513\\[-1pt]{\fontsize{6}{7}\selectfont(0.057)}} \\
Protein & \shortstack[r]{\underline{1.745}\\[-1pt]{\fontsize{6}{7}\selectfont(0.018)}} & \shortstack[r]{\underline{1.738}\\[-1pt]{\fontsize{6}{7}\selectfont(0.017)}} & \shortstack[r]{\textbf{1.729}\\[-1pt]{\fontsize{6}{7}\selectfont(0.014)}} & \shortstack[r]{2.022\\[-1pt]{\fontsize{6}{7}\selectfont(0.075)}} & \shortstack[r]{1.848\\[-1pt]{\fontsize{6}{7}\selectfont(0.024)}} & \shortstack[r]{1.993\\[-1pt]{\fontsize{6}{7}\selectfont(0.028)}} & \shortstack[r]{2.303\\[-1pt]{\fontsize{6}{7}\selectfont(0.024)}} & \shortstack[r]{2.142\\[-1pt]{\fontsize{6}{7}\selectfont(0.026)}} & \shortstack[r]{2.016\\[-1pt]{\fontsize{6}{7}\selectfont(0.027)}} & \shortstack[r]{2.700\\[-1pt]{\fontsize{6}{7}\selectfont(0.024)}} \\
Power plant & \shortstack[r]{\underline{1.646}\\[-1pt]{\fontsize{6}{7}\selectfont(0.021)}} & \shortstack[r]{\underline{1.673}\\[-1pt]{\fontsize{6}{7}\selectfont(0.026)}} & \shortstack[r]{\textbf{1.534}\\[-1pt]{\fontsize{6}{7}\selectfont(0.062)}} & \shortstack[r]{\underline{1.960}\\[-1pt]{\fontsize{6}{7}\selectfont(0.185)}} & \shortstack[r]{\underline{1.629}\\[-1pt]{\fontsize{6}{7}\selectfont(0.023)}} & \shortstack[r]{1.753\\[-1pt]{\fontsize{6}{7}\selectfont(0.032)}} & \shortstack[r]{2.149\\[-1pt]{\fontsize{6}{7}\selectfont(0.038)}} & \shortstack[r]{2.109\\[-1pt]{\fontsize{6}{7}\selectfont(0.038)}} & \shortstack[r]{2.024\\[-1pt]{\fontsize{6}{7}\selectfont(0.052)}} & \shortstack[r]{2.353\\[-1pt]{\fontsize{6}{7}\selectfont(0.019)}} \\
\midrule Mean rank & \textbf{2.1} & 3.4 & \underline{2.3} & 6.3 & \underline{3.8} & 5.1 & 7.8 & 9.0 & 6.9 & 8.3 \\
\bottomrule
\end{tabular*}
\end{table*}

The ranking depends on which aspect of the distribution is scored.
FPT-D and FPT have the best mean NLL ranks (1.8 and 2.2).
They achieve lower mean NLL than the best non-FPT comparator on seven
and six of the ten datasets, respectively (\cref{tab:reg-logloss}).
Conditional Tree Flow leads on PCE with mean rank 1.1, versus 4.9 for
FPT-D and 6.6 for FPT.
See \cref{tab:univariate-main} in \cref{app:results-real} for the
cross-metric summary.

The recorded prediction phase averages
13.7 seconds per test fold for FPT, 169.3 for Treeffuser and 0.04 for
CatBoost-R (\cref{fig:real-predict}). Tuning FPT reduces mean CRPS by
a median of 1.6\% relative to FPT-D. Search and refitting together take
1,049 seconds per fold on average, compared with 40.7 seconds to fit
FPT-D. \Cref{tab:univariate-timing} in \cref{app:results-real} reports
prediction and metric evaluation times separately.

FPT is competitive with Treeffuser and CatBoost, avoids CatBoost's
Gaussian assumption, and trains and predicts faster than Treeffuser.

\begin{table*}[!t]
\centering
\scriptsize
\setlength{\tabcolsep}{2pt}
\renewcommand{\arraystretch}{1.18}
\caption{FPT ablations relative to the full model on ten datasets. Each metric shows its mean within-dataset rank (lower is better) and the median score change from FPT-D. Bold marks the best mean rank; underlined ranks are not significantly different from it (two-sided Wilcoxon signed-rank test over settings \citep{demsar2006statistical,benavoli2016should}, on log score ratios except for NLL differences, Holm-adjusted, $\alpha=0.05$). CRPS changes are relative percentages; NLL and PCE changes are absolute. Positive changes indicate worse scores. Single anchor retains only the exhaustive member. Per-dataset means and sample SDs appear in \Cref{sec:ablation-results}.}
\label{tab:flow-ablation}
\begin{tabular*}{\linewidth}{@{\extracolsep{\fill}}lrrrrrr@{}}
\toprule
Configuration & CRPS rank & $\Delta$CRPS & NLL rank & $\Delta$NLL & PCE rank & $\Delta$PCE \\
\midrule
Full model (FPT-D) & \textbf{1.0} & --- & \textbf{1.6} & --- & 3.6 & --- \\
Without random-pair candidates & 2.6 & +2.42\% & 2.6 & +0.051 & \underline{2.8} & -0.0012 \\
Without target smoothing & 2.4 & +1.19\% & \underline{2.1} & +0.008 & \textbf{1.8} & -0.0070 \\
Single anchor & 4.0 & +9.91\% & 4.2 & +0.537 & 4.3 & +0.0089 \\
Single anchor, without target smoothing & 5.0 & +11.94\% & \underline{4.5} & +0.758 & \underline{2.5} & -0.0070 \\
\bottomrule
\end{tabular*}
\end{table*}

\begin{figure*}[!t]
\centering
\includegraphics[width=\textwidth]{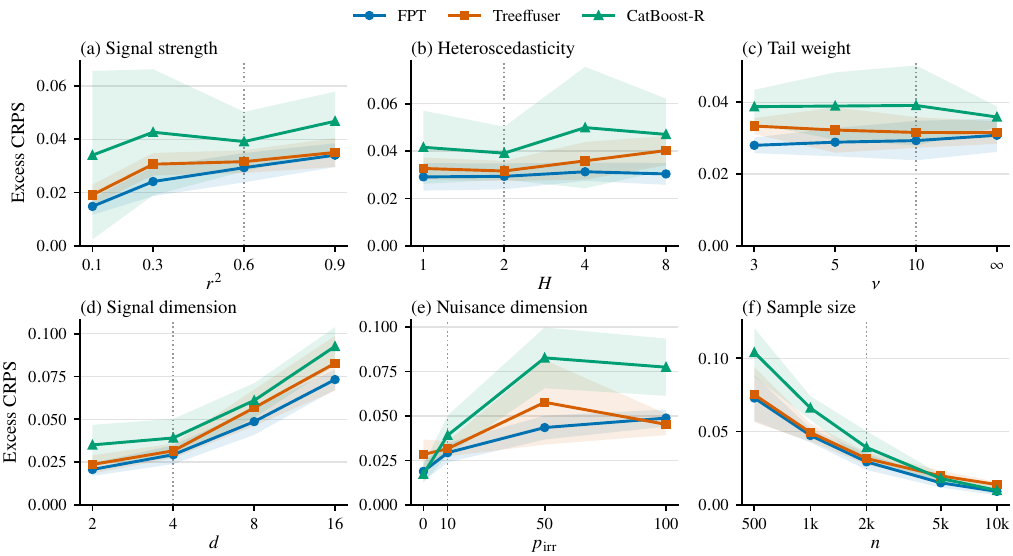}
\caption{Mean excess CRPS for three CRPS-tuned models as one factor varies. Dotted lines mark the common reference with $r^2=0.6$, $H=2$, Student-$t$ noise with 10 degrees of freedom, four relevant and ten irrelevant features, and 2,000 training observations. Shading spans one sample SD over five replicates. CatBoost-R uses RMSE early stopping.}
\label{fig:syn-sweeps}
\end{figure*}

\subsection{Contributions of the ensemble and training choices}
\label{sec:exp-ablation}

\paragraph{Datasets and evaluation.}
We use the same ten regression datasets and shuffled five-fold
cross-validation splits as in \cref{sec:exp-real}, and evaluate
configurations using CRPS, NLL and PCE.

\paragraph{Training procedure.}
The ablations use the FPT-D hyperparameters. Each fit selects its own stopping round
by validation CRPS.

\paragraph{Results.}
To isolate the effect of averaging with row subsampling, we compare one subsampled path that
searches target splits exhaustively (the \emph{anchor}) with an ensemble
of nine such paths. Averaging reduces mean CRPS on all ten datasets,
by a median of 5.25\% relative to the anchor. We then retain one exhaustive path and 
enable random-pair split points on the other 8 paths. This mixed
ensemble (FPT-D) has lower mean CRPS than the exhaustive ensemble on
all ten datasets, by a median of 2.36\%. In summary, ensembling
and random-pair averaging contribute both to the performance of FPT.


Enabling smoothing reduces mean CRPS on all ten
datasets, both for the mixed ensemble (median reduction 1.18\%) and for the anchor
alone (3.14\%), relative to treating responses as points. On the other hand, smoothing
increases decile calibration error on nine datasets for the
ensemble and all ten for the anchor. 
\Cref{tab:flow-ablation} compares each configuration with FPT-D, and
\cref{sec:ablation-results} gives the scores on each dataset.

\subsection{Sensitivity to the conditional distribution}
\label{sec:exp-synthetic}

\paragraph{Datasets and models.}
We compare tuned FPT, Treeffuser and CatBoost-R, and separately
compare FPT, Treeffuser and CatBoost with fixed defaults, on 28
synthetic settings with known conditional distributions. We vary
signal strength, heteroscedasticity, tail weight, relevant and irrelevant
feature counts, and training size in isolation (\cref{fig:syn-sweeps}).
\Cref{app:results-synthetic} gives the data-generating process and settings.

\paragraph{Training and test procedure.}
All methods share five independent training/test replicates per setting,
with 2,000 fresh test observations each and seeds reused across settings.
Tuning minimizes validation CRPS on an 80/20 training split, up to
100 trials or 20 minutes, checked between trials. FPT stops by
CRPS and CatBoost-R by validation RMSE.

\paragraph{Results.}
In \Cref{fig:syn-sweeps}, tuned FPT has the lowest mean excess CRPS
in 26 of 28 settings. The two exceptions are both on the irrelevant-feature
sweep, where CatBoost-R leads with zero irrelevant features and Treeffuser
with 100. The win count conceals how close FPT and Treeffuser often are.
FPT's CRPS is lower by a median of 1.1\% relative to Treeffuser, and the
paired 95\% intervals for their excess-CRPS difference include zero in
22 settings. On the training-size sweep, increasing the sample from 500
to 10,000 reduces FPT's mean excess CRPS from 0.0729 to 0.0091.

FPT and Treeffuser both cover 87.4\% of test responses on average with
their nominal 90\% intervals, versus 69.2\% for CatBoost-R
(\cref{fig:syn-calibration}).
\Cref{fig:syn-quartiles} shows how coverage varies with the noise scale.

\section{Conclusion}
\label{sec:conclusion}

Flow Partition Trees combine joint-space partitioning with conditional
CDF boosting to estimate flexible predictive distributions with directly
evaluable densities, CDFs and quantiles. The population partition score
lower bounds per-round log-loss improvement after normalization and
shrinkage. Posterior fractional counts extend the construction to
interval likelihoods, with a corresponding improvement bound.

On ten real-data datasets, FPT achieves the best average CRPS rank and
its default configuration the best average NLL rank. Its CRPS is competitive
with Treeffuser, with lower recorded training and prediction times.
Ablations support the benefits of path averaging, random-pair splits
and target smoothing.

The guarantees concern population trees, and geometric convergence requires
a weak-learning assumption. The implementation uses bounded response
support and an adaptive bin width that can exceed measurement precision;
better CRPS does not consistently yield better calibration. Extending the
analysis to fitted trees and understanding the interaction between bin
width, density accuracy and calibration remain open directions.

\bibliography{references}
\bibliographystyle{icml2027/icml2027}

\clearpage
\appendix
\onecolumn

\let\addcontentsline\appendixaddcontentsline
\setcounter{tocdepth}{2}
\renewcommand{\contentsname}{Appendix contents}
\tableofcontents
\clearpage

\section{Proofs}
\label{app:proofs}

\subsection{Bregman decomposition}
\label{app:criteria}

{\LinkTargetOff\bregmandecomposition*}
\begin{proof}
Apply the three-point identity of Bregman divergence twice:
\begin{align*}
  B_\phi(f,g_\pi) &= B_\phi(f, f_\pi) + B_\phi(f_\pi, g_\pi)
  + \int(\phi'(f_\pi) - \phi'(g_\pi))(f - f_\pi)d\mathbb{Q} \\
  B_\phi(f,1) &= B_\phi(f,f_\pi) + B_\phi(f_\pi, 1) + \int (\phi'(f_\pi) - \phi'(1))(f - f_\pi) d\mathbb{Q}
\end{align*}

On each leaf $A$, the first cross term contributes
\[
 (\phi'(c_A)-\phi'(g_A))\int_A(f-c_A)\,d\QQ
 = (\phi'(c_A)-\phi'(g_A))(p_A-q_Ac_A)=0.
\]
The same calculation with $g_A=1$ eliminates the second cross term.
Subtracting the two identities and using
$B_\phi(f_\pi,g_\pi)=\sum_A q_A d_\phi(c_A,g_A)$ gives
\eqref{eq:Bregman-decomposition}. Strict convexity makes each summand
nonnegative, with equality exactly when $g_A=c_A$.
\end{proof}

\subsection{Per-round KL improvement}
\label{app:accounting}

\perroundimprovement*
\begin{proof}
Using the density update \eqref{eq:fpt} and substituting $u=\Psi_{t-1}(y\mid X)$, we have
\begin{align*}
  \calK_{t-1}-\calK_t
  &= \EE_{\PX}\!\left[\int_{I_0} f\left(\log\frac{f}{\psi_{t-1}}-\log\frac{f}{\psi_t}\right)\,dy\right] \\
  &= \EE_{\PX}\!\left[\int_{I_0} f\log\frac{\psi_t}{\psi_{t-1}}\,dy\right] \\
  &= \EE_{\PX}\!\left[\int_{I_0} f\log\tilde f_{\pi_t}\bigl(\Psi_{t-1}(y\mid X)\mid X\bigr)\,dy\right] \\
  &= \EE_{\PX}\!\left[\int_0^1 r_{t-1}(X,u)\log\tilde f_{\pi_t}(u\mid X)\,du\right] \\
  &= \EE_{\PP_{t-1}}\log\tilde f_{\pi_t}(U^{(t-1)}\mid X).
\end{align*}

By Jensen's inequality,
\begin{align*}
\log\tilde f_{\pi_t} &\geq (1-\alpha)\log 1+\alpha\log\bar f_{\pi_t} \\
&= \alpha\log f_{\pi_t}-\alpha\log\varsigma_t(X).
\end{align*}
Since $\EE_{\PX}\varsigma_t(X)=\int f_{\pi_t}\,d\QQ_u=\sum_{A\in\pi_t}p_A^{(t-1)}=1$, another application of Jensen's inequality gives
\begin{align*}
  \EE_{\PP_{t-1}}[\log\varsigma_t(X)]
  &= \int \log\varsigma_t(x)\underbrace{\left(\int_0^1 r_{t-1}(x,u)\,du\right)}_{=1}\PX(dx) \\
  &= \EE_{\PX}[\log\varsigma_t(X)] \\
  &\leq \log\EE_{\PX}[\varsigma_t(X)]=0.
\end{align*}
Taking expectations in the pointwise bound and using
$\EE_{\PP_{t-1}}\log f_{\pi_t}=D_t(\pi_t)$ by \eqref{eq:divergence-step-t},
we obtain $\calK_{t-1}-\calK_t\geq\alpha D_t(\pi_t)\geq 0$.
\end{proof}

\geometricdecay*
\begin{proof}
Under the stated hypothesis, the bound gives
$\calK_t\leq(1-\alpha\gamma)\calK_{t-1}$. Iterating over the $T$ rounds
proves \eqref{eq:kl-decay}.
\end{proof}

\subsection{Bin likelihood and uniform dequantization}
\label{app:quantized}

\begin{corollary}[Lower bound on the bin log-likelihood gain]
\label{cor:bin-improvement}
Suppose $I_0$ contains the observed bins, $\psi_{t-1}>0$ on $I_0$, and
$L_\delta(\psi_{t-1})$ is finite. Let $\tilde f_{\pi_t}$ be the positive
layer obtained by normalizing the tree with E-step leaf values $p_A^{(t-1)}/q_A$,
$q_A>0$, and shrinking it with $\alpha\in[0,1]$. Then
\[
 L_\delta(\psi_t)-L_\delta(\psi_{t-1})
 \geq \EE_{\PP_{t-1,\delta}}\log\tilde f_{\pi_t}(U^{(t-1)}\mid X)
 \geq \alpha D_t(\pi_t)\geq0.
\]
\end{corollary}
\begin{proof}
Integrating \eqref{eq:fpt} over a bin $B$ at $x$ and changing variables gives
\begin{align*}
  m_B^{(t)}(x) = \int_B \psi_t(y \mid x) dy &= \int_{I_B^{(t-1)}}\tilde{f}_{\pi_t}(u\mid x)du
\end{align*}
And
\begin{equation*}
 \frac{m_B^{(t)}(x)}{m_B^{(t-1)}(x)}
 =\frac{1}{\mu(I_B^{(t-1)}(x))}
   \int_{I_B^{(t-1)}(x)} \tilde f_{\pi_t}(u\mid x)\,du.
\end{equation*}
By Jensen's inequality:
\begin{align*}
L_\delta(\psi_t) - L_\delta(\psi_{t-1}) &= \EE_{(X, \bar{Y})}
[\log \frac{m_{B(\bar{Y})}^{(t)}(X)}{m_{B(\bar{Y})}^{(t-1)}(X)}] \\
&= \EE_{(X, \bar{Y})} \left[ \log \left( \frac{1}{\mu(I_{B(\bar{Y})}^{(t-1)}(X))}
   \int_{I_{B(\bar{Y})}^{(t-1)}(X)} \tilde f_{\pi_t}(u\mid X)\,du \right) \right] \\
&\geq \EE_{(X, \bar{Y})} \left[ \frac{1}{\mu(I_{B(\bar{Y})}^{(t-1)}(X))}
   \int_{I_{B(\bar{Y})}^{(t-1)}(X)} \log  \tilde f_{\pi_t}(u\mid X)\,du \right] \\
&= \EE_{\PP_{t-1, \delta}}[\log \tilde{f}_{\pi_t}(U^{(t-1)} \mid X)].
\end{align*}

This gives the first inequality.

Under the E-step law the raw tree has expected log value
$\sum_A p_A^{(t-1)}\log(p_A^{(t-1)}/q_A)=D_t(\pi_t)$, and its normalizer satisfies
$\EE_{\PX}\varsigma_t(X)=\sum_A p_A^{(t-1)}=1$. The normalization and shrinkage
argument in \eqref{eq:bregman-flow-progress} therefore bounds the
expected log layer multiplier below by $\alpha D_t(\pi_t)$.
\end{proof}

The gain in bin log-likelihood also equals the decrease in quantized KL.
Indeed, for the true bin probabilities $p_B(X)$ and fitted bin probabilities
$m_B^{(t)}(X)$,
\[
 \calK_{t,\delta}
 =\EE_{\PX}\sum_{B:\,p_B(X)>0}p_B(X)\log\frac{p_B(X)}{m_B^{(t)}(X)},
 \qquad
 \calK_{t-1,\delta}-\calK_{t,\delta}=L_\delta(\psi_t)-L_\delta(\psi_{t-1}),
\]
since the true bin probabilities cancel in the log ratio.




\section{Fitting and evaluating FPT}
\label{app:implementation}

\subsection{Automatic target bin width}
\label{app:bin-width}

With \texttt{target\_smoothing}$=\texttt{auto}$, we estimate a bin width
$\delta$ from the responses supplied to each fit; Optuna does not tune
this width. We first infer a measurement spacing $h$ such that roughly
95\% or more of the responses are, within numerical tolerance, integer
multiples of $h$. For example, $1.2$ and $1.3$ match spacing $h=0.1$,
whereas $1.27$ does not. All responses remain in the fit. We initialize
$\delta=h$; if spacing inference fails, we use the smallest positive gap
between distinct responses. We enlarge this initial width as needed
to keep the response support, including its 0.2 expansion factor, within
$2^{16}$ bins, to avoid memory issues in our implementation.

For a candidate width, let $b_i(\delta)=\operatorname{round}(y_i/\delta)$
and $N_k(\delta)=\sum_{i=1}^n\1_{\{b_i(\delta)=k\}}$. We coarsen the
width until
\[
 \frac1n\sum_{i=1}^n
 \1_{\{N_{b_i(\delta)}(\delta)\geq2\}}\geq0.95.
\]
Thus at least 95\% of observations share their bin with another
observation. When coarsening is needed, we bracket a satisfying width by
doubling and then perform up to 24 refinement steps. When a spacing $h$
is inferred, candidate widths remain integer multiples of $h$. We round the
responses to $\bar y_i=\delta\,b_i(\delta)$ and fit the corresponding
intervals $[\bar y_i-\delta/2,\bar y_i+\delta/2)$ using the procedure in
\cref{sec:quantized}. The width is shared by all flow paths within a fit
and is recomputed for each tuning fit and outer refit. It can therefore
differ across folds and can be coarser than the measurement resolution.

\subsection{Computing conditional quantiles}
\label{app:quantile-evaluation}

For $\tau \in (0,1)$, define the conditional quantile after
round $t$ as
\begin{equation}
\label{eq:quantile}
  q_t(\tau \mid x)
  = \inf\{y : \Psi_t(y \mid x) \ge \tau\}.
\end{equation}
Quantiles are computed by reversing the sequence of transports.
Starting from $v_t=\tau$, we recursively evaluate
\[
  v_{s-1} = \tilde F_{\pi_s}^{-1}(v_s \mid x),
  \qquad s=t,\ldots,1,
\]
and return $q_t(\tau \mid x)=F_0^{-1}(v_0)$.
Each layer CDF is piecewise linear, so inversion consists of
locating the corresponding cumulative-probability interval
and interpolating within it. With the uniform initialization,
$F_0^{-1}(v)=a_0+(b_0-a_0)v$.

During validation, we cache compositions of consecutive rounds
in blocks whose lengths are distinct powers of two. Adding a
round merges adjacent blocks of equal length. Quantile queries
invert these cached maps from newest to oldest, requiring at
most $\lfloor\log_2 t\rfloor+1$ block inversions instead of
inverting each round separately.

\subsection{Early stopping by validation CRPS}
\label{sec:early-stopping}

The improvement bound in \cref{sec:bregman} concerns the population
objective. To select a stopping round by CRPS, we score each fitted
path on held-out responses after every round.
For a residual $r=y-q_t(\tau\mid x)$, the pinball loss is
\[
  \rho_\tau(r)
  = \bigl(\tau-\1_{\{r<0\}}\bigr)r.
\]
The quantile representation of CRPS gives
\[
  \operatorname{CRPS}(\Psi_t(\cdot\mid x),y)
  = 2\int_0^1
    \rho_\tau\bigl(y-q_t(\tau\mid x)\bigr)\,d\tau.
\]
We approximate this integral using the midpoint rule with
$K=10$ levels, $\tau_j=(j-\tfrac12)/K$:
\[
  \widehat{\operatorname{CRPS}}_{K,t}(x,y)
  = \frac{2}{K}\sum_{j=1}^{K}
    \rho_{\tau_j}\bigl(y-q_t(\tau_j\mid x)\bigr).
\]
Early stopping monitors the average validation score.
The ten quantiles are computed from the fitted path as in \cref{app:quantile-evaluation}.

\section{Model selection and benchmark configurations}
\label{app:hyperparameters}

\subsection{Shared tuning protocol}
\label{app:tuning}

For each tuned method, we minimize validation CRPS using Optuna's
tree-structured Parzen estimator. Each outer training fold uses one
shuffled 80/20 tuning split and a fresh search, limited to 100 trials
or 20 minutes. The time limit is checked between trials, so a running
trial can exceed it. We refit the selected configuration on the outer
training fold. Methods with early stopping reserve a separate validation
split within each fit (\cref{app:fpt-configurations,app:baseline-configurations}).
The five-fold real-data split, tuning split and search sampler
use seed 42. The synthetic tuned comparisons use the same search spaces
and budgets, separately for each setting and replicate.

In the search spaces, $[a,b]_{\log}$ denotes a logarithmic range and
$[a,b]$ denotes a linear range. Counts and depths take integer
values, and braces denote categorical choices. Parameter names omit
preprocessing pipeline prefixes.

Treeffuser, NGBoost, LinCDE and Conditional Tree Flow receive dense
one-hot encodings of categorical covariates, fitted within each
training split; the partition methods and CatBoost use their native
categorical handling. Available parallel-worker settings are set to
use all cores in the real-data benchmark and tuned synthetic runs.

\subsection{FPT configurations}
\label{app:fpt-configurations}

FPT and FPT-D each average nine flow paths. One uses exhaustive target
split candidates, and the other eight each use up to four random-pair
candidates per node. Each path samples 80\% of the training rows without
replacement and uses a feature fraction of 0.8. We use at most 1,000
rounds, a 20\% internal validation split and patience 30 while fitting
the NLL objective. We monitor validation CRPS (\cref{sec:early-stopping})
in the real-data experiments and tuned synthetic runs. \Cref{app:results-synthetic}
gives the stopping rule for the fixed-default synthetic comparison.
The base CDF is affine, the support-expansion factor is 0.2, and we select
the target bin width as in \cref{app:bin-width}.
Covariate splits use histograms with at most 256 bins. Other fixed
controls are \texttt{leaf\_smoothing\_lambda}$=1$,
\texttt{min\_gain}$=10^{-9}$,
\texttt{min\_samples\_xy\_x\_ratio}$=0$,
\texttt{min\_samples\_y}$=0$, and \texttt{balanced\_tree}$=\mathrm{False}$.
\Cref{tab:fpt-hyperparameters} gives the five parameters varied by the
tuned method and their fixed FPT-D values.

\begin{table}[!htbp]
\centering
\small
\caption{FPT search space and fixed FPT-D values. All five search ranges
use logarithmic scaling.}
\label{tab:fpt-hyperparameters}
\begin{tabular}{@{}lrr@{}}
\toprule
Parameter & FPT-D & FPT search range \\
\midrule
\texttt{learning\_rate} & 0.3 & $[0.03,0.5]_{\log}$ \\
\texttt{max\_leaves} & 32 & $[4,128]_{\log}$ \\
\texttt{min\_samples\_split} & 50 & $[1,200]_{\log}$ \\
\texttt{min\_samples\_x\_split\_ratio} & 0.01 & $[0.001,1]_{\log}$ \\
\texttt{min\_volume\_fraction} & 0.01 & $[0.001,0.4]_{\log}$ \\
\bottomrule
\end{tabular}
\end{table}

\subsection{Baseline configurations}
\label{app:baseline-configurations}

\paragraph{Partition Tree and Partition Forest.}
These controls use unboosted density trees, histogram covariate splits
with at most 256 bins and support-expansion factor 0.2. Their common
tree search space and the forest's additional ensemble parameters are
listed in \cref{tab:partition-hyperparameters}. Neither uses boosting
early stopping.

\begin{table}[!htbp]
\centering
\small
\caption{Search spaces for Partition Tree (PT) and Partition Forest (PF).
The forest varies the same eight tree parameters as PT and three
additional ensemble parameters.}
\label{tab:partition-hyperparameters}
\begin{tabular}{@{}lll@{}}
\toprule
Parameter & Search range & Methods \\
\midrule
\texttt{max\_leaves} & $[8,1024]_{\log}$ & PT, PF \\
\texttt{max\_depth} & $[3,25]_{\log}$ & PT, PF \\
\texttt{min\_samples\_split} & $[2,200]_{\log}$ & PT, PF \\
\texttt{min\_samples\_xy} & $[1,100]_{\log}$ & PT, PF \\
\texttt{min\_samples\_x} & $[1,100]_{\log}$ & PT, PF \\
\texttt{min\_samples\_y} & $[1,100]_{\log}$ & PT, PF \\
\texttt{min\_volume\_fraction} & $[0,0.4]$ & PT, PF \\
\texttt{min\_gain} & $[10^{-13},10^{-2}]_{\log}$ & PT, PF \\
\addlinespace
\texttt{n\_estimators} & $[25,300]_{\log}$ & PF \\
\texttt{max\_samples} & $[0.5,1]$ & PF \\
\texttt{max\_features} & $[0.3,1]$ & PF \\
\bottomrule
\end{tabular}
\end{table}

\Cref{tab:baseline-hyperparameters} gives the fitting and prediction settings
alongside the search space for each remaining baseline.

\begin{table}[H]
\centering
\small
\caption{Configurations and search spaces for the tuned baselines.}
\label{tab:baseline-hyperparameters}
\begin{tabular}{@{}p{0.48\linewidth}@{\hspace{0.03\linewidth}}p{0.49\linewidth}@{}}
\toprule
Model and fixed settings & Tuned parameters and search domains \\
\midrule
\raggedright \textbf{CatBoost and CatBoost-R.}
Both fit Gaussian predictive distributions with the \texttt{RMSEWithUncertainty}
objective, Bernoulli bootstrapping, plain boosting and \texttt{max\_bin}$=254$.
Each reserves 10\% of the fitting data for early stopping, with patience 200,
separately from the tuning holdout. CatBoost monitors NLL for up to
5,000 iterations. CatBoost-R monitors RMSE for up to 20,000 iterations.
& \raggedright \emph{CatBoost.} \texttt{learning\_rate}: $[0.005,0.05]_{\log}$;
\texttt{depth}: $[4,8]$;
\texttt{l2\_leaf\_reg}: $[0.05,5]_{\log}$.\par\smallskip
\emph{CatBoost-R.} \texttt{learning\_rate}: $[0.005,0.2]_{\log}$;
\texttt{depth}: $[4,10]$;
\texttt{l2\_leaf\_reg}: $[0.05,20]_{\log}$. \tabularnewline
\addlinespace
\raggedright \textbf{Treeffuser.}
The LightGBM score model uses 10\% internal validation, at most 255
histogram bins, at least 20 observations per leaf, and no depth cap.
The variance-exploding SDE uses its default scales and 50 Euler steps
per predictive trajectory. We take 100 draws per observation in batches
of ten on real data and 256 draws in one batch on synthetic data.
For NLL we use a Gaussian kernel density estimate with Scott's
bandwidth rule and a minimum bandwidth of $10^{-6}$.
& \raggedright \texttt{n\_estimators}: $[100,3000]$;
\texttt{n\_repeats}: $[10,50]$;
\texttt{learning\_rate}: $[0.01,1]_{\log}$;
\texttt{early\_stopping\_rounds}: $[10,100]$;
\texttt{num\_leaves}: $[10,100]$. \tabularnewline
\addlinespace
\raggedright \textbf{NGBoost.}
We use a Normal predictive family, the logarithmic score and natural
gradients. The weak learner is a regression tree with the
\texttt{friedman\_mse} criterion. Early stopping uses a 20\% validation
split and patience 200. The \texttt{estimator\_\_} prefix denotes tree
parameters. The value \texttt{None} uses all features for
\texttt{max\_features} and leaves the leaf count unrestricted for
\texttt{max\_leaf\_nodes}.
& \raggedright \texttt{n\_estimators}: $[200,10000]_{\log}$;
\texttt{learning\_rate}: $[0.005,0.1]_{\log}$;
\texttt{minibatch\_frac}: $[0.5,1]$;
\texttt{col\_sample}: $[0.5,1]$;
\texttt{estimator\_\_max\_depth}: $[2,8]$;
\texttt{estimator\_\_min\_samples\_split}: $[2,40]_{\log}$;
\texttt{estimator\_\_min\_samples\_leaf}: $[1,20]_{\log}$;
\texttt{estimator\_\_max\_features}: $\{\mathrm{None},\texttt{sqrt},\texttt{log2}\}$;
\texttt{estimator\_\_max\_leaf\_nodes}: $\{\mathrm{None},8,16,32,64,128\}$. \tabularnewline
\addlinespace
\raggedright \textbf{LinCDE.}
We use the boosted R implementation through its probabilistic-regression
adapter, with the \texttt{nsTransform} spline basis, a Gaussian prior,
response-grid expansion 0.1 and centering disabled. The CRPS search
selects the number of boosting trees.
& \raggedright \texttt{split\_point}: $[10,50]$;
\texttt{spline\_df}: $[5,15]$;
\texttt{number\_bin}: $[20,80]$;
\texttt{df}: $[2,8]$;
\texttt{depth}: $[1,5]$;
\texttt{terminal\_size}: $[10,100]_{\log}$;
\texttt{alpha}: $[0.05,0.5]$;
\texttt{subsample}: $[0.5,1]$;
\texttt{n\_trees}: $[25,250]_{\log}$;
\texttt{shrinkage}: $[0.01,0.3]_{\log}$. \tabularnewline
\addlinespace
\raggedright \textbf{Conditional Tree Flow.}
Each node uses standardized features and logistic regression with the
\texttt{lbfgs} solver and at most 1,000 solver iterations. We use the
logistic outcome transform, at most 300 flow trees, split-location penalty
\texttt{eta}$=0.1$, a 10\% internal validation fraction and a likelihood-based
stopping window of ten rounds. $C$ is the node classifier's inverse
regularization strength.
& \raggedright \texttt{max\_depth}: $[2,5]$;
\texttt{min\_samples\_split}: $[10,100]_{\log}$;
\texttt{n\_cutpoints}: $\{3,5,9\}$;
\texttt{c0}: $[0.03,0.3]_{\log}$;
\texttt{gamma}: $[0,1]$;
$C$: $[0.01,100]_{\log}$. \tabularnewline
\bottomrule
\end{tabular}
\end{table}

\clearpage

\section{Experimental details and additional results}
\label{app:results}

\subsection{Predictive performance on real data}
\label{app:results-real}

\begin{table}[!htbp]
\centering
\small
\setlength{\tabcolsep}{2pt}
\renewcommand{\arraystretch}{1.18}
\caption{Cross-metric summary over ten datasets: rank columns average within-dataset ranks (lower is better); CRPS wins count the lowest dataset means. Bold marks the best mean rank; underlined ranks are not significantly different from it (two-sided Wilcoxon signed-rank test over settings \citep{demsar2006statistical,benavoli2016should}, on log score ratios except for NLL differences, Holm-adjusted, $\alpha=0.05$). PT, PF and CTF denote Partition Tree, Partition Forest and Conditional Tree Flow. FPT is tuned; FPT-D uses fixed hyperparameters. CatBoost stops boosting on validation NLL, CatBoost-R on validation RMSE with a wider search space.}
\label{tab:univariate-main}
\begin{tabular*}{\linewidth}{@{\extracolsep{\fill}}lrrrrr@{}}
\toprule
Model & CRPS rank & CRPS wins & NLL rank & RMSE rank & PCE rank \\
\midrule
FPT & \textbf{2.1} & 3 & \underline{2.2} & \underline{2.6} & 6.6 \\
FPT-D & 3.4 & 0 & \textbf{1.8} & 4.0 & 4.9 \\
Treeffuser & \underline{2.3} & 4 & 5.2 & \underline{2.9} & 3.4 \\
CatBoost & 6.3 & 0 & 4.6 & 6.1 & 5.6 \\
CatBoost-R & \underline{3.8} & 2 & 9.4 & \textbf{2.0} & 9.0 \\
NGBoost & 5.1 & 0 & 6.9 & 4.4 & 7.6 \\
LinCDE & 7.8 & 0 & 5.9 & 7.7 & 6.4 \\
PT & 9.0 & 0 & 7.7 & 9.1 & 3.3 \\
PF & 6.9 & 0 & 5.0 & 7.4 & 7.1 \\
CTF & 8.3 & 1 & 6.3 & 8.8 & \textbf{1.1} \\
\bottomrule
\end{tabular*}
\end{table}

\begin{figure}[!htbp]
\centering
\includegraphics[width=\textwidth]{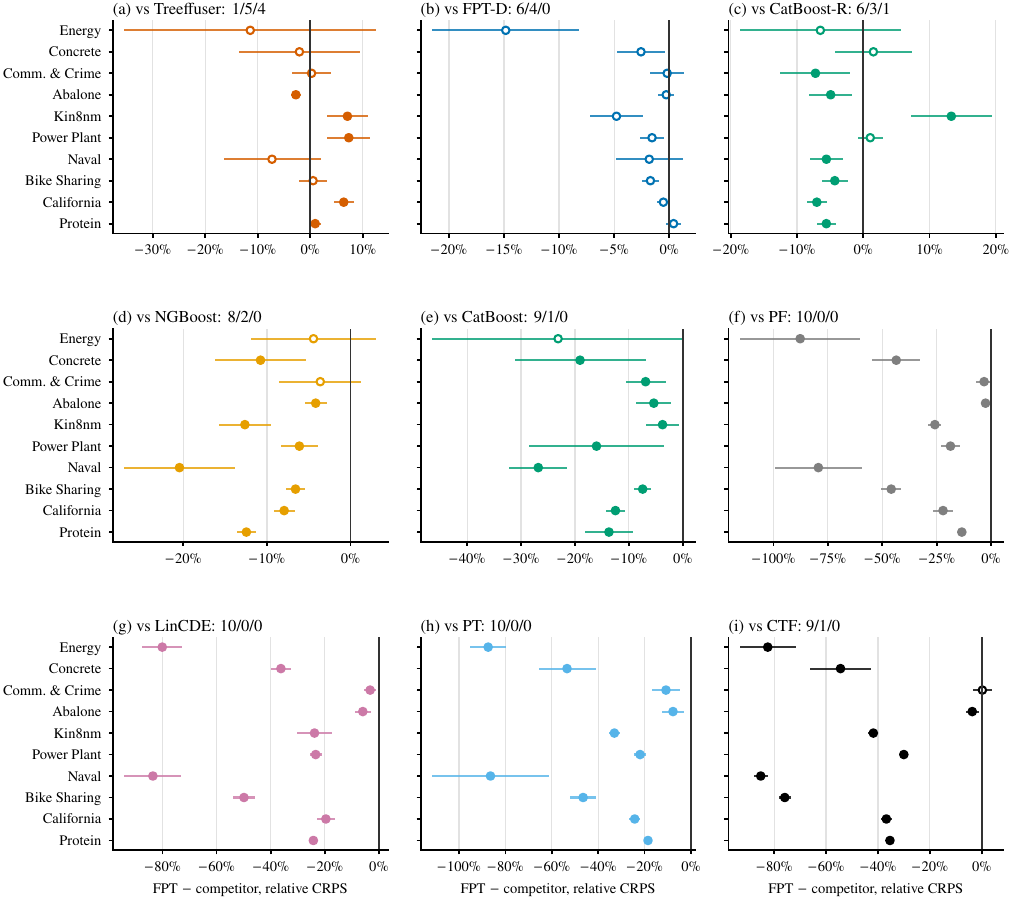}
\caption{Dataset-mean CRPS difference between FPT and each comparator, relative to the comparator. Negative values favor FPT. Bars give unadjusted 95\% paired $t$ intervals over five folds. Panel titles count intervals below zero, containing zero and above zero. Overlap between training folds makes these intervals exploratory. Each panel has its own horizontal scale.}
\label{fig:real-baselines-crps}
\end{figure}

\Cref{fig:real-baselines-crps} shows
the CRPS differences on each dataset, and \cref{tab:reg-logloss,tab:reg-rmse}
give the corresponding NLL and RMSE scores.

\begin{table*}[!t]
\centering
\scriptsize
\setlength{\tabcolsep}{2pt}
\renewcommand{\arraystretch}{1.18}
\caption{Negative log-likelihood: five-fold means, with sample SDs beneath in parentheses. A multiplier beside a dataset applies to both means and SDs in that row. Bold marks the lowest unrounded mean; underlined means are not significantly different from it (two-sided corrected resampled $t$-test over the five folds \citep{nadeau2003inference}, Holm-adjusted within the row \citep{holm1979simple}, $\alpha=0.05$). The final row averages within-dataset ranks across the ten datasets (lower is better); its marks follow \Cref{tab:univariate-main}. TF denotes Treeffuser, CB CatBoost, CB-R CatBoost-R, and NG NGBoost; remaining abbreviations follow \Cref{tab:univariate-main}.}
\label{tab:reg-logloss}
\begin{tabular*}{\linewidth}{@{\extracolsep{\fill}}lrrrrrrrrrr@{}}
\toprule
Dataset & FPT & FPT-D & TF & CB & CB-R & NG & LinCDE & PT & PF & CTF \\
\midrule
Abalone & \shortstack[r]{\underline{1.967}\\[-1pt]{\fontsize{6}{7}\selectfont(0.049)}} & \shortstack[r]{\underline{1.971}\\[-1pt]{\fontsize{6}{7}\selectfont(0.051)}} & \shortstack[r]{\underline{2.056}\\[-1pt]{\fontsize{6}{7}\selectfont(0.077)}} & \shortstack[r]{\underline{2.103}\\[-1pt]{\fontsize{6}{7}\selectfont(0.059)}} & \shortstack[r]{\underline{2.275}\\[-1pt]{\fontsize{6}{7}\selectfont(0.287)}} & \shortstack[r]{\underline{2.115}\\[-1pt]{\fontsize{6}{7}\selectfont(0.079)}} & \shortstack[r]{\underline{2.020}\\[-1pt]{\fontsize{6}{7}\selectfont(0.058)}} & \shortstack[r]{\underline{2.140}\\[-1pt]{\fontsize{6}{7}\selectfont(0.080)}} & \shortstack[r]{\underline{1.986}\\[-1pt]{\fontsize{6}{7}\selectfont(0.045)}} & \shortstack[r]{\textbf{1.724}\\[-1pt]{\fontsize{6}{7}\selectfont(0.332)}} \\
Bike sharing & \shortstack[r]{\underline{4.515}\\[-1pt]{\fontsize{6}{7}\selectfont(0.091)}} & \shortstack[r]{\textbf{4.465}\\[-1pt]{\fontsize{6}{7}\selectfont(0.036)}} & \shortstack[r]{\underline{4.719}\\[-1pt]{\fontsize{6}{7}\selectfont(0.171)}} & \shortstack[r]{4.711\\[-1pt]{\fontsize{6}{7}\selectfont(0.062)}} & \shortstack[r]{4.910\\[-1pt]{\fontsize{6}{7}\selectfont(0.123)}} & \shortstack[r]{4.726\\[-1pt]{\fontsize{6}{7}\selectfont(0.071)}} & \shortstack[r]{5.525\\[-1pt]{\fontsize{6}{7}\selectfont(0.092)}} & \shortstack[r]{5.261\\[-1pt]{\fontsize{6}{7}\selectfont(0.056)}} & \shortstack[r]{5.166\\[-1pt]{\fontsize{6}{7}\selectfont(0.055)}} & \shortstack[r]{5.849\\[-1pt]{\fontsize{6}{7}\selectfont(0.012)}} \\
California & \shortstack[r]{\textbf{0.091}\\[-1pt]{\fontsize{6}{7}\selectfont(0.049)}} & \shortstack[r]{\underline{0.097}\\[-1pt]{\fontsize{6}{7}\selectfont(0.046)}} & \shortstack[r]{\underline{0.768}\\[-1pt]{\fontsize{6}{7}\selectfont(0.410)}} & \shortstack[r]{0.524\\[-1pt]{\fontsize{6}{7}\selectfont(0.041)}} & \shortstack[r]{0.677\\[-1pt]{\fontsize{6}{7}\selectfont(0.146)}} & \shortstack[r]{0.600\\[-1pt]{\fontsize{6}{7}\selectfont(0.054)}} & \shortstack[r]{0.546\\[-1pt]{\fontsize{6}{7}\selectfont(0.025)}} & \shortstack[r]{0.666\\[-1pt]{\fontsize{6}{7}\selectfont(0.030)}} & \shortstack[r]{0.507\\[-1pt]{\fontsize{6}{7}\selectfont(0.066)}} & \shortstack[r]{0.736\\[-1pt]{\fontsize{6}{7}\selectfont(0.016)}} \\
Communities & \shortstack[r]{\textbf{-1.009}\\[-1pt]{\fontsize{6}{7}\selectfont(0.063)}} & \shortstack[r]{\underline{-0.993}\\[-1pt]{\fontsize{6}{7}\selectfont(0.072)}} & \shortstack[r]{\underline{-0.752}\\[-1pt]{\fontsize{6}{7}\selectfont(0.055)}} & \shortstack[r]{\underline{-0.697}\\[-1pt]{\fontsize{6}{7}\selectfont(0.264)}} & \shortstack[r]{\underline{1.345}\\[-1pt]{\fontsize{6}{7}\selectfont(3.109)}} & \shortstack[r]{\underline{-0.615}\\[-1pt]{\fontsize{6}{7}\selectfont(0.305)}} & \shortstack[r]{\underline{-0.891}\\[-1pt]{\fontsize{6}{7}\selectfont(0.073)}} & \shortstack[r]{-0.733\\[-1pt]{\fontsize{6}{7}\selectfont(0.050)}} & \shortstack[r]{\underline{-0.933}\\[-1pt]{\fontsize{6}{7}\selectfont(0.078)}} & \shortstack[r]{\underline{-0.990}\\[-1pt]{\fontsize{6}{7}\selectfont(0.063)}} \\
Concrete & \shortstack[r]{\underline{2.778}\\[-1pt]{\fontsize{6}{7}\selectfont(0.212)}} & \shortstack[r]{\textbf{2.735}\\[-1pt]{\fontsize{6}{7}\selectfont(0.139)}} & \shortstack[r]{\underline{3.295}\\[-1pt]{\fontsize{6}{7}\selectfont(0.442)}} & \shortstack[r]{\underline{3.180}\\[-1pt]{\fontsize{6}{7}\selectfont(0.186)}} & \shortstack[r]{\underline{6.345}\\[-1pt]{\fontsize{6}{7}\selectfont(2.548)}} & \shortstack[r]{\underline{4.788}\\[-1pt]{\fontsize{6}{7}\selectfont(2.191)}} & \shortstack[r]{3.302\\[-1pt]{\fontsize{6}{7}\selectfont(0.042)}} & \shortstack[r]{3.708\\[-1pt]{\fontsize{6}{7}\selectfont(0.095)}} & \shortstack[r]{3.359\\[-1pt]{\fontsize{6}{7}\selectfont(0.137)}} & \shortstack[r]{3.510\\[-1pt]{\fontsize{6}{7}\selectfont(0.095)}} \\
Energy {\tiny($10^{2}$)} & \shortstack[r]{\textbf{0.0028}\\[-1pt]{\fontsize{6}{7}\selectfont(0.0012)}} & \shortstack[r]{\underline{0.0035}\\[-1pt]{\fontsize{6}{7}\selectfont(0.0009)}} & \shortstack[r]{\underline{0.0146}\\[-1pt]{\fontsize{6}{7}\selectfont(0.0187)}} & \shortstack[r]{\underline{0.0102}\\[-1pt]{\fontsize{6}{7}\selectfont(0.0079)}} & \shortstack[r]{\underline{1.1741}\\[-1pt]{\fontsize{6}{7}\selectfont(2.0031)}} & \shortstack[r]{\underline{0.0069}\\[-1pt]{\fontsize{6}{7}\selectfont(0.0043)}} & \shortstack[r]{0.0198\\[-1pt]{\fontsize{6}{7}\selectfont(0.0005)}} & \shortstack[r]{0.0222\\[-1pt]{\fontsize{6}{7}\selectfont(0.0025)}} & \shortstack[r]{0.0218\\[-1pt]{\fontsize{6}{7}\selectfont(0.0027)}} & \shortstack[r]{0.0150\\[-1pt]{\fontsize{6}{7}\selectfont(0.0007)}} \\
Kin8nm & \shortstack[r]{-0.437\\[-1pt]{\fontsize{6}{7}\selectfont(0.062)}} & \shortstack[r]{-0.661\\[-1pt]{\fontsize{6}{7}\selectfont(0.024)}} & \shortstack[r]{\underline{-0.662}\\[-1pt]{\fontsize{6}{7}\selectfont(0.091)}} & \shortstack[r]{\textbf{-0.708}\\[-1pt]{\fontsize{6}{7}\selectfont(0.030)}} & \shortstack[r]{\underline{9.042}\\[-1pt]{\fontsize{6}{7}\selectfont(12.278)}} & \shortstack[r]{-0.350\\[-1pt]{\fontsize{6}{7}\selectfont(0.111)}} & \shortstack[r]{-0.493\\[-1pt]{\fontsize{6}{7}\selectfont(0.040)}} & \shortstack[r]{-0.147\\[-1pt]{\fontsize{6}{7}\selectfont(0.037)}} & \shortstack[r]{-0.431\\[-1pt]{\fontsize{6}{7}\selectfont(0.028)}} & \shortstack[r]{-0.236\\[-1pt]{\fontsize{6}{7}\selectfont(0.018)}} \\
Naval & \shortstack[r]{\underline{-5.747}\\[-1pt]{\fontsize{6}{7}\selectfont(0.041)}} & \shortstack[r]{\underline{-5.782}\\[-1pt]{\fontsize{6}{7}\selectfont(0.015)}} & \shortstack[r]{\textbf{-5.848}\\[-1pt]{\fontsize{6}{7}\selectfont(0.111)}} & \shortstack[r]{\underline{-5.672}\\[-1pt]{\fontsize{6}{7}\selectfont(0.035)}} & \shortstack[r]{\underline{0.184}\\[-1pt]{\fontsize{6}{7}\selectfont(7.824)}} & \shortstack[r]{-5.092\\[-1pt]{\fontsize{6}{7}\selectfont(0.124)}} & \shortstack[r]{-4.033\\[-1pt]{\fontsize{6}{7}\selectfont(0.065)}} & \shortstack[r]{-3.818\\[-1pt]{\fontsize{6}{7}\selectfont(0.222)}} & \shortstack[r]{-4.236\\[-1pt]{\fontsize{6}{7}\selectfont(0.158)}} & \shortstack[r]{-4.449\\[-1pt]{\fontsize{6}{7}\selectfont(0.041)}} \\
Protein & \shortstack[r]{2.211\\[-1pt]{\fontsize{6}{7}\selectfont(0.019)}} & \shortstack[r]{\textbf{1.883}\\[-1pt]{\fontsize{6}{7}\selectfont(0.018)}} & \shortstack[r]{2.297\\[-1pt]{\fontsize{6}{7}\selectfont(0.102)}} & \shortstack[r]{2.685\\[-1pt]{\fontsize{6}{7}\selectfont(0.044)}} & \shortstack[r]{3.318\\[-1pt]{\fontsize{6}{7}\selectfont(0.506)}} & \shortstack[r]{3.037\\[-1pt]{\fontsize{6}{7}\selectfont(0.216)}} & \shortstack[r]{2.471\\[-1pt]{\fontsize{6}{7}\selectfont(0.011)}} & \shortstack[r]{2.278\\[-1pt]{\fontsize{6}{7}\selectfont(0.056)}} & \shortstack[r]{2.210\\[-1pt]{\fontsize{6}{7}\selectfont(0.088)}} & \shortstack[r]{2.583\\[-1pt]{\fontsize{6}{7}\selectfont(0.010)}} \\
Power plant & \shortstack[r]{\underline{2.660}\\[-1pt]{\fontsize{6}{7}\selectfont(0.080)}} & \shortstack[r]{\textbf{2.528}\\[-1pt]{\fontsize{6}{7}\selectfont(0.020)}} & \shortstack[r]{\underline{3.340}\\[-1pt]{\fontsize{6}{7}\selectfont(0.329)}} & \shortstack[r]{\underline{2.753}\\[-1pt]{\fontsize{6}{7}\selectfont(0.082)}} & \shortstack[r]{\underline{3.578}\\[-1pt]{\fontsize{6}{7}\selectfont(0.759)}} & \shortstack[r]{\underline{2.884}\\[-1pt]{\fontsize{6}{7}\selectfont(0.204)}} & \shortstack[r]{2.780\\[-1pt]{\fontsize{6}{7}\selectfont(0.022)}} & \shortstack[r]{2.881\\[-1pt]{\fontsize{6}{7}\selectfont(0.032)}} & \shortstack[r]{\underline{2.660}\\[-1pt]{\fontsize{6}{7}\selectfont(0.033)}} & \shortstack[r]{2.827\\[-1pt]{\fontsize{6}{7}\selectfont(0.007)}} \\
\midrule Mean rank & \underline{2.2} & \textbf{1.8} & 5.2 & 4.6 & 9.4 & 6.9 & 5.9 & 7.7 & 5.0 & 6.3 \\
\bottomrule
\end{tabular*}
\end{table*}

\begin{table*}[!t]
\centering
\scriptsize
\setlength{\tabcolsep}{2pt}
\renewcommand{\arraystretch}{1.18}
\caption{RMSE: five-fold means, with sample SDs beneath in parentheses. A multiplier beside a dataset applies to both means and SDs in that row. Bold marks the lowest unrounded mean; underlined means are not significantly different from it (two-sided corrected resampled $t$-test over the five folds \citep{nadeau2003inference}, Holm-adjusted within the row \citep{holm1979simple}, $\alpha=0.05$). The final row averages within-dataset ranks across the ten datasets (lower is better); its marks follow \Cref{tab:univariate-main}. TF denotes Treeffuser, CB CatBoost, CB-R CatBoost-R, and NG NGBoost; remaining abbreviations follow \Cref{tab:univariate-main}.}
\label{tab:reg-rmse}
\begin{tabular*}{\linewidth}{@{\extracolsep{\fill}}lrrrrrrrrrr@{}}
\toprule
Dataset & FPT & FPT-D & TF & CB & CB-R & NG & LinCDE & PT & PF & CTF \\
\midrule
Abalone & \shortstack[r]{\underline{2.137}\\[-1pt]{\fontsize{6}{7}\selectfont(0.148)}} & \shortstack[r]{\textbf{2.136}\\[-1pt]{\fontsize{6}{7}\selectfont(0.148)}} & \shortstack[r]{\underline{2.181}\\[-1pt]{\fontsize{6}{7}\selectfont(0.154)}} & \shortstack[r]{\underline{2.178}\\[-1pt]{\fontsize{6}{7}\selectfont(0.123)}} & \shortstack[r]{\underline{2.150}\\[-1pt]{\fontsize{6}{7}\selectfont(0.145)}} & \shortstack[r]{\underline{2.158}\\[-1pt]{\fontsize{6}{7}\selectfont(0.150)}} & \shortstack[r]{\underline{2.220}\\[-1pt]{\fontsize{6}{7}\selectfont(0.158)}} & \shortstack[r]{\underline{2.247}\\[-1pt]{\fontsize{6}{7}\selectfont(0.171)}} & \shortstack[r]{\underline{2.154}\\[-1pt]{\fontsize{6}{7}\selectfont(0.156)}} & \shortstack[r]{\underline{2.195}\\[-1pt]{\fontsize{6}{7}\selectfont(0.170)}} \\
Bike sharing {\tiny($10^{2}$)} & \shortstack[r]{\underline{0.387}\\[-1pt]{\fontsize{6}{7}\selectfont(0.010)}} & \shortstack[r]{0.394\\[-1pt]{\fontsize{6}{7}\selectfont(0.009)}} & \shortstack[r]{\underline{0.376}\\[-1pt]{\fontsize{6}{7}\selectfont(0.009)}} & \shortstack[r]{0.396\\[-1pt]{\fontsize{6}{7}\selectfont(0.010)}} & \shortstack[r]{\textbf{0.375}\\[-1pt]{\fontsize{6}{7}\selectfont(0.013)}} & \shortstack[r]{0.395\\[-1pt]{\fontsize{6}{7}\selectfont(0.015)}} & \shortstack[r]{0.625\\[-1pt]{\fontsize{6}{7}\selectfont(0.028)}} & \shortstack[r]{0.686\\[-1pt]{\fontsize{6}{7}\selectfont(0.029)}} & \shortstack[r]{0.677\\[-1pt]{\fontsize{6}{7}\selectfont(0.024)}} & \shortstack[r]{1.383\\[-1pt]{\fontsize{6}{7}\selectfont(0.024)}} \\
California & \shortstack[r]{0.453\\[-1pt]{\fontsize{6}{7}\selectfont(0.011)}} & \shortstack[r]{0.454\\[-1pt]{\fontsize{6}{7}\selectfont(0.013)}} & \shortstack[r]{\textbf{0.428}\\[-1pt]{\fontsize{6}{7}\selectfont(0.011)}} & \shortstack[r]{0.484\\[-1pt]{\fontsize{6}{7}\selectfont(0.016)}} & \shortstack[r]{0.456\\[-1pt]{\fontsize{6}{7}\selectfont(0.010)}} & \shortstack[r]{0.460\\[-1pt]{\fontsize{6}{7}\selectfont(0.011)}} & \shortstack[r]{0.527\\[-1pt]{\fontsize{6}{7}\selectfont(0.011)}} & \shortstack[r]{0.560\\[-1pt]{\fontsize{6}{7}\selectfont(0.010)}} & \shortstack[r]{0.550\\[-1pt]{\fontsize{6}{7}\selectfont(0.024)}} & \shortstack[r]{0.697\\[-1pt]{\fontsize{6}{7}\selectfont(0.077)}} \\
Communities & \shortstack[r]{\underline{0.1360}\\[-1pt]{\fontsize{6}{7}\selectfont(0.0121)}} & \shortstack[r]{\underline{0.1363}\\[-1pt]{\fontsize{6}{7}\selectfont(0.0134)}} & \shortstack[r]{\underline{0.1349}\\[-1pt]{\fontsize{6}{7}\selectfont(0.0124)}} & \shortstack[r]{\underline{0.1397}\\[-1pt]{\fontsize{6}{7}\selectfont(0.0149)}} & \shortstack[r]{\underline{0.1350}\\[-1pt]{\fontsize{6}{7}\selectfont(0.0111)}} & \shortstack[r]{\textbf{0.1343}\\[-1pt]{\fontsize{6}{7}\selectfont(0.0124)}} & \shortstack[r]{\underline{0.1385}\\[-1pt]{\fontsize{6}{7}\selectfont(0.0119)}} & \shortstack[r]{\underline{0.1481}\\[-1pt]{\fontsize{6}{7}\selectfont(0.0179)}} & \shortstack[r]{\underline{0.1394}\\[-1pt]{\fontsize{6}{7}\selectfont(0.0132)}} & \shortstack[r]{\underline{0.1345}\\[-1pt]{\fontsize{6}{7}\selectfont(0.0124)}} \\
Concrete & \shortstack[r]{\underline{4.469}\\[-1pt]{\fontsize{6}{7}\selectfont(0.539)}} & \shortstack[r]{\underline{4.599}\\[-1pt]{\fontsize{6}{7}\selectfont(0.433)}} & \shortstack[r]{\underline{4.611}\\[-1pt]{\fontsize{6}{7}\selectfont(0.408)}} & \shortstack[r]{\underline{5.116}\\[-1pt]{\fontsize{6}{7}\selectfont(0.349)}} & \shortstack[r]{\textbf{4.156}\\[-1pt]{\fontsize{6}{7}\selectfont(0.477)}} & \shortstack[r]{4.670\\[-1pt]{\fontsize{6}{7}\selectfont(0.595)}} & \shortstack[r]{6.258\\[-1pt]{\fontsize{6}{7}\selectfont(0.668)}} & \shortstack[r]{8.776\\[-1pt]{\fontsize{6}{7}\selectfont(0.770)}} & \shortstack[r]{7.210\\[-1pt]{\fontsize{6}{7}\selectfont(1.018)}} & \shortstack[r]{9.362\\[-1pt]{\fontsize{6}{7}\selectfont(0.895)}} \\
Energy & \shortstack[r]{\textbf{0.389}\\[-1pt]{\fontsize{6}{7}\selectfont(0.055)}} & \shortstack[r]{\underline{0.471}\\[-1pt]{\fontsize{6}{7}\selectfont(0.051)}} & \shortstack[r]{\underline{0.464}\\[-1pt]{\fontsize{6}{7}\selectfont(0.096)}} & \shortstack[r]{\underline{0.503}\\[-1pt]{\fontsize{6}{7}\selectfont(0.106)}} & \shortstack[r]{\underline{0.393}\\[-1pt]{\fontsize{6}{7}\selectfont(0.037)}} & \shortstack[r]{\underline{0.404}\\[-1pt]{\fontsize{6}{7}\selectfont(0.053)}} & \shortstack[r]{1.916\\[-1pt]{\fontsize{6}{7}\selectfont(0.170)}} & \shortstack[r]{3.405\\[-1pt]{\fontsize{6}{7}\selectfont(0.270)}} & \shortstack[r]{3.366\\[-1pt]{\fontsize{6}{7}\selectfont(0.770)}} & \shortstack[r]{2.616\\[-1pt]{\fontsize{6}{7}\selectfont(0.301)}} \\
Kin8nm & \shortstack[r]{0.119\\[-1pt]{\fontsize{6}{7}\selectfont(0.004)}} & \shortstack[r]{0.128\\[-1pt]{\fontsize{6}{7}\selectfont(0.004)}} & \shortstack[r]{0.108\\[-1pt]{\fontsize{6}{7}\selectfont(0.003)}} & \shortstack[r]{0.122\\[-1pt]{\fontsize{6}{7}\selectfont(0.003)}} & \shortstack[r]{\textbf{0.090}\\[-1pt]{\fontsize{6}{7}\selectfont(0.003)}} & \shortstack[r]{0.129\\[-1pt]{\fontsize{6}{7}\selectfont(0.004)}} & \shortstack[r]{0.156\\[-1pt]{\fontsize{6}{7}\selectfont(0.008)}} & \shortstack[r]{0.174\\[-1pt]{\fontsize{6}{7}\selectfont(0.005)}} & \shortstack[r]{0.160\\[-1pt]{\fontsize{6}{7}\selectfont(0.003)}} & \shortstack[r]{0.201\\[-1pt]{\fontsize{6}{7}\selectfont(0.004)}} \\
Naval {\tiny($10^{-3}$)} & \shortstack[r]{0.811\\[-1pt]{\fontsize{6}{7}\selectfont(0.036)}} & \shortstack[r]{0.823\\[-1pt]{\fontsize{6}{7}\selectfont(0.041)}} & \shortstack[r]{\underline{0.767}\\[-1pt]{\fontsize{6}{7}\selectfont(0.024)}} & \shortstack[r]{1.121\\[-1pt]{\fontsize{6}{7}\selectfont(0.086)}} & \shortstack[r]{\textbf{0.742}\\[-1pt]{\fontsize{6}{7}\selectfont(0.030)}} & \shortstack[r]{0.939\\[-1pt]{\fontsize{6}{7}\selectfont(0.069)}} & \shortstack[r]{4.291\\[-1pt]{\fontsize{6}{7}\selectfont(0.388)}} & \shortstack[r]{4.696\\[-1pt]{\fontsize{6}{7}\selectfont(0.573)}} & \shortstack[r]{3.370\\[-1pt]{\fontsize{6}{7}\selectfont(0.546)}} & \shortstack[r]{7.653\\[-1pt]{\fontsize{6}{7}\selectfont(0.457)}} \\
Protein & \shortstack[r]{3.693\\[-1pt]{\fontsize{6}{7}\selectfont(0.027)}} & \shortstack[r]{3.833\\[-1pt]{\fontsize{6}{7}\selectfont(0.022)}} & \shortstack[r]{3.813\\[-1pt]{\fontsize{6}{7}\selectfont(0.025)}} & \shortstack[r]{3.896\\[-1pt]{\fontsize{6}{7}\selectfont(0.110)}} & \shortstack[r]{\textbf{3.562}\\[-1pt]{\fontsize{6}{7}\selectfont(0.039)}} & \shortstack[r]{3.820\\[-1pt]{\fontsize{6}{7}\selectfont(0.053)}} & \shortstack[r]{4.645\\[-1pt]{\fontsize{6}{7}\selectfont(0.038)}} & \shortstack[r]{4.477\\[-1pt]{\fontsize{6}{7}\selectfont(0.060)}} & \shortstack[r]{4.265\\[-1pt]{\fontsize{6}{7}\selectfont(0.068)}} & \shortstack[r]{5.137\\[-1pt]{\fontsize{6}{7}\selectfont(0.035)}} \\
Power plant & \shortstack[r]{\underline{3.238}\\[-1pt]{\fontsize{6}{7}\selectfont(0.100)}} & \shortstack[r]{\underline{3.295}\\[-1pt]{\fontsize{6}{7}\selectfont(0.094)}} & \shortstack[r]{\textbf{3.023}\\[-1pt]{\fontsize{6}{7}\selectfont(0.187)}} & \shortstack[r]{\underline{3.692}\\[-1pt]{\fontsize{6}{7}\selectfont(0.254)}} & \shortstack[r]{\underline{3.095}\\[-1pt]{\fontsize{6}{7}\selectfont(0.082)}} & \shortstack[r]{\underline{3.316}\\[-1pt]{\fontsize{6}{7}\selectfont(0.086)}} & \shortstack[r]{3.966\\[-1pt]{\fontsize{6}{7}\selectfont(0.111)}} & \shortstack[r]{3.954\\[-1pt]{\fontsize{6}{7}\selectfont(0.097)}} & \shortstack[r]{3.851\\[-1pt]{\fontsize{6}{7}\selectfont(0.103)}} & \shortstack[r]{4.302\\[-1pt]{\fontsize{6}{7}\selectfont(0.069)}} \\
\midrule Mean rank & \underline{2.6} & 4.0 & \underline{2.9} & 6.1 & \textbf{2.0} & 4.4 & 7.7 & 9.1 & 7.4 & 8.8 \\
\bottomrule
\end{tabular*}
\end{table*}

\begin{figure}[!htbp]
\centering
\includegraphics[width=\textwidth]{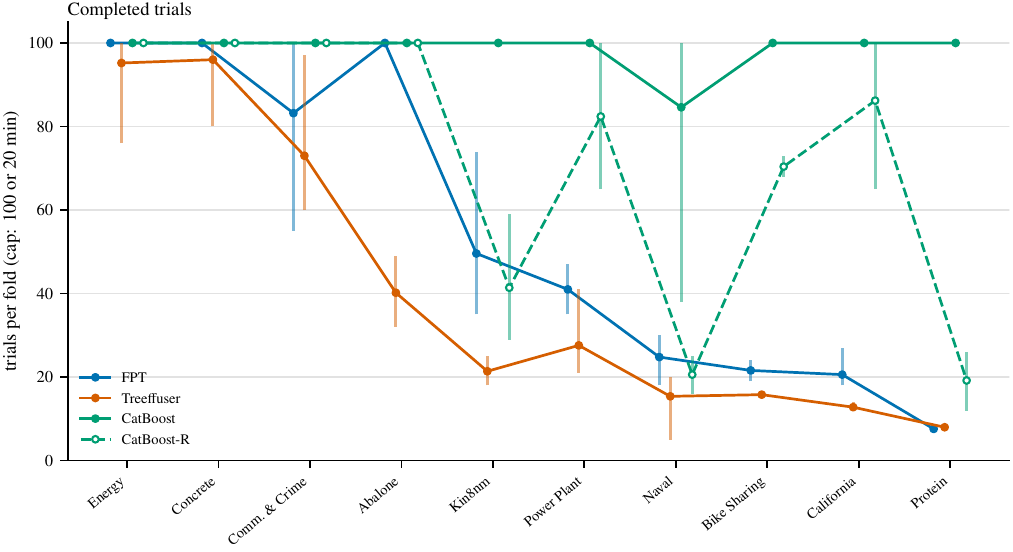}
\caption{Completed Optuna trials per dataset (mean and min--max over folds) under the shared budget of at most 100 trials or 20 minutes per fold. Budget and dataset size are confounded.}
\label{fig:real-budget}
\end{figure}

\begin{figure}[!htbp]
\centering
\includegraphics[width=\textwidth]{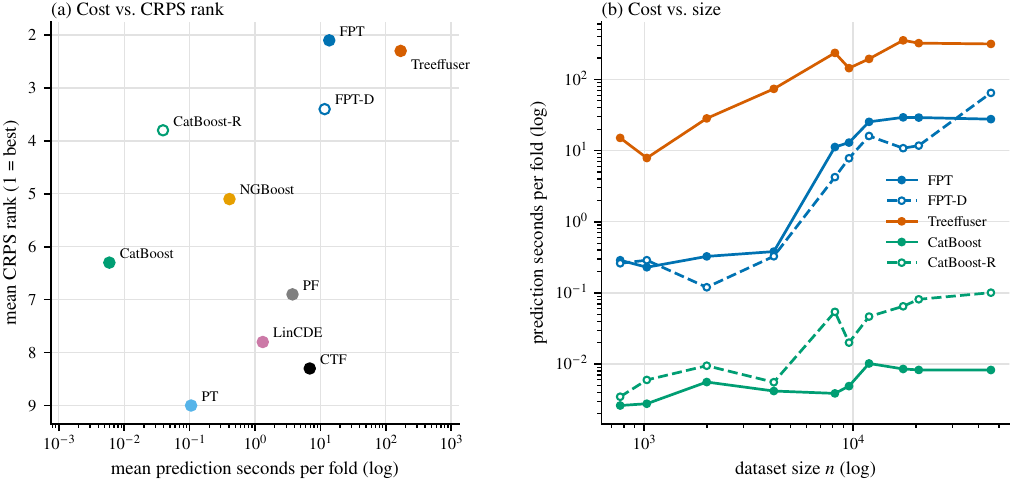}
\caption{Time taken to predict an outer test fold, averaged over the five folds of each dataset. The left panel compares mean prediction time and mean CRPS rank over datasets. The right panel shows how prediction time varies with dataset size.}
\label{fig:real-predict}
\end{figure}

\FloatBarrier
\paragraph{Computational cost.}
\Cref{fig:real-budget} reports how many trials each search completed.
\Cref{fig:real-predict} relates prediction time to CRPS rank and dataset
size. \Cref{tab:univariate-timing} separates search and refit time from
prediction and metric evaluation.

\begin{table*}[!t]
\centering
\small
\setlength{\tabcolsep}{2pt}
\renewcommand{\arraystretch}{1.18}
\caption{Mean recorded seconds per real-data outer fold, averaged over all 50 folds. Fit includes search and refit for tuned methods; FPT-D performs one fit. Prediction and metric evaluation are separate recorded phases. These are pipeline timings, not isolated model-inference benchmarks.}
\label{tab:univariate-timing}
\begin{tabular*}{\linewidth}{@{\extracolsep{\fill}}lrrr@{}}
\toprule
Model & Fit/search (s) & Predict (s) & Metrics (s) \\
\midrule
FPT & 1,048.8 & 13.7 & 0.6119 \\
FPT-D & 40.7 & 11.6 & 0.9234 \\
Treeffuser & 1,212.9 & 169.3 & 0.0653 \\
CatBoost & 174.6 & 0.0059 & 0.0041 \\
CatBoost-R & 895.9 & 0.0392 & 0.0065 \\
NGBoost & 674.2 & 0.4074 & 0.0041 \\
LinCDE & 970.2 & 1.3 & 15.6 \\
PT & 30.9 & 0.1054 & 0.1363 \\
PF & 374.5 & 3.8 & 3.8 \\
CTF & 1,656.0 & 6.9 & 10.2 \\
\bottomrule
\end{tabular*}
\end{table*}

\FloatBarrier
\subsection{Contributions of the ensemble and training choices}
\label{sec:ablation-results}

Each configuration uses the fixed FPT-D values in \cref{tab:fpt-hyperparameters},
samples 80\% of the rows per path, and uses a feature fraction of 0.8.
The anchor is one path with exhaustive target splits. The exhaustive
ensemble averages nine such paths. The mixed ensemble (FPT-D) uses one
exhaustive path and eight random-pair paths. For the unsmoothed versions
of the anchor and the mixed ensemble, we set \texttt{target\_smoothing}
to \texttt{None}. Each fit selects its stopping round by validation CRPS.

\Cref{tab:ablation-crps,tab:ablation-logloss,tab:ablation-pce1} report
scores for these five configurations, and \cref{tab:ablation-runtime}
reports their timings. The separate all-random-pair comparison uses
random-pair target splits on all nine paths.

\begin{table*}[!t]
\centering
\scriptsize
\setlength{\tabcolsep}{2pt}
\renewcommand{\arraystretch}{1.18}
\caption{CRPS for the five complete FPT configurations: means over five folds, with sample SDs beneath. Row multipliers apply to means and SDs. Bold marks the lowest unrounded mean; underlined means are not significantly different from it (two-sided corrected resampled $t$-test over the five folds \citep{nadeau2003inference}, Holm-adjusted within the row \citep{holm1979simple}, $\alpha=0.05$). The final row averages ranks among these five configurations; its marks follow \Cref{tab:flow-ablation}. Smooth denotes automatic target smoothing.}
\label{tab:ablation-crps}
\begin{tabular*}{\linewidth}{@{\extracolsep{\fill}}lrrrrr@{}}
\toprule
Dataset & Mixed + smooth & Exhaustive + smooth & Anchor + smooth & Mixed & Anchor \\
\midrule
Abalone & \shortstack[r]{\textbf{1.055}\\[-1pt]{\fontsize{6}{7}\selectfont(0.062)}} & \shortstack[r]{\underline{1.056}\\[-1pt]{\fontsize{6}{7}\selectfont(0.065)}} & \shortstack[r]{1.078\\[-1pt]{\fontsize{6}{7}\selectfont(0.060)}} & \shortstack[r]{1.069\\[-1pt]{\fontsize{6}{7}\selectfont(0.059)}} & \shortstack[r]{1.092\\[-1pt]{\fontsize{6}{7}\selectfont(0.062)}} \\
Bike sharing & \shortstack[r]{\textbf{16.69}\\[-1pt]{\fontsize{6}{7}\selectfont(0.61)}} & \shortstack[r]{17.07\\[-1pt]{\fontsize{6}{7}\selectfont(0.48)}} & \shortstack[r]{18.35\\[-1pt]{\fontsize{6}{7}\selectfont(0.50)}} & \shortstack[r]{17.04\\[-1pt]{\fontsize{6}{7}\selectfont(0.56)}} & \shortstack[r]{19.39\\[-1pt]{\fontsize{6}{7}\selectfont(0.38)}} \\
California & \shortstack[r]{\textbf{0.208}\\[-1pt]{\fontsize{6}{7}\selectfont(0.006)}} & \shortstack[r]{\underline{0.210}\\[-1pt]{\fontsize{6}{7}\selectfont(0.004)}} & \shortstack[r]{0.218\\[-1pt]{\fontsize{6}{7}\selectfont(0.005)}} & \shortstack[r]{\underline{0.209}\\[-1pt]{\fontsize{6}{7}\selectfont(0.005)}} & \shortstack[r]{0.223\\[-1pt]{\fontsize{6}{7}\selectfont(0.004)}} \\
Communities & \shortstack[r]{\textbf{0.064}\\[-1pt]{\fontsize{6}{7}\selectfont(0.006)}} & \shortstack[r]{\underline{0.065}\\[-1pt]{\fontsize{6}{7}\selectfont(0.007)}} & \shortstack[r]{\underline{0.067}\\[-1pt]{\fontsize{6}{7}\selectfont(0.007)}} & \shortstack[r]{\underline{0.065}\\[-1pt]{\fontsize{6}{7}\selectfont(0.006)}} & \shortstack[r]{0.069\\[-1pt]{\fontsize{6}{7}\selectfont(0.006)}} \\
Concrete & \shortstack[r]{\textbf{2.296}\\[-1pt]{\fontsize{6}{7}\selectfont(0.211)}} & \shortstack[r]{2.360\\[-1pt]{\fontsize{6}{7}\selectfont(0.214)}} & \shortstack[r]{2.698\\[-1pt]{\fontsize{6}{7}\selectfont(0.259)}} & \shortstack[r]{\underline{2.413}\\[-1pt]{\fontsize{6}{7}\selectfont(0.278)}} & \shortstack[r]{2.891\\[-1pt]{\fontsize{6}{7}\selectfont(0.357)}} \\
Energy & \shortstack[r]{\textbf{0.233}\\[-1pt]{\fontsize{6}{7}\selectfont(0.013)}} & \shortstack[r]{\underline{0.251}\\[-1pt]{\fontsize{6}{7}\selectfont(0.016)}} & \shortstack[r]{\underline{0.281}\\[-1pt]{\fontsize{6}{7}\selectfont(0.022)}} & \shortstack[r]{\underline{0.237}\\[-1pt]{\fontsize{6}{7}\selectfont(0.019)}} & \shortstack[r]{\underline{0.321}\\[-1pt]{\fontsize{6}{7}\selectfont(0.053)}} \\
Kin8nm & \shortstack[r]{\textbf{0.068}\\[-1pt]{\fontsize{6}{7}\selectfont(0.002)}} & \shortstack[r]{0.070\\[-1pt]{\fontsize{6}{7}\selectfont(0.003)}} & \shortstack[r]{0.075\\[-1pt]{\fontsize{6}{7}\selectfont(0.002)}} & \shortstack[r]{\underline{0.069}\\[-1pt]{\fontsize{6}{7}\selectfont(0.002)}} & \shortstack[r]{0.077\\[-1pt]{\fontsize{6}{7}\selectfont(0.003)}} \\
Naval {\tiny($10^{-4}$)} & \shortstack[r]{\textbf{3.786}\\[-1pt]{\fontsize{6}{7}\selectfont(0.106)}} & \shortstack[r]{\underline{3.856}\\[-1pt]{\fontsize{6}{7}\selectfont(0.095)}} & \shortstack[r]{5.586\\[-1pt]{\fontsize{6}{7}\selectfont(0.102)}} & \shortstack[r]{4.414\\[-1pt]{\fontsize{6}{7}\selectfont(0.111)}} & \shortstack[r]{7.221\\[-1pt]{\fontsize{6}{7}\selectfont(0.322)}} \\
Protein & \shortstack[r]{\textbf{1.738}\\[-1pt]{\fontsize{6}{7}\selectfont(0.017)}} & \shortstack[r]{1.868\\[-1pt]{\fontsize{6}{7}\selectfont(0.009)}} & \shortstack[r]{1.916\\[-1pt]{\fontsize{6}{7}\selectfont(0.023)}} & \shortstack[r]{\underline{1.754}\\[-1pt]{\fontsize{6}{7}\selectfont(0.019)}} & \shortstack[r]{1.945\\[-1pt]{\fontsize{6}{7}\selectfont(0.015)}} \\
Power plant & \shortstack[r]{\textbf{1.673}\\[-1pt]{\fontsize{6}{7}\selectfont(0.026)}} & \shortstack[r]{1.715\\[-1pt]{\fontsize{6}{7}\selectfont(0.030)}} & \shortstack[r]{1.789\\[-1pt]{\fontsize{6}{7}\selectfont(0.044)}} & \shortstack[r]{\underline{1.691}\\[-1pt]{\fontsize{6}{7}\selectfont(0.026)}} & \shortstack[r]{1.864\\[-1pt]{\fontsize{6}{7}\selectfont(0.032)}} \\
\midrule Mean rank & \textbf{1.0} & 2.6 & 4.0 & 2.4 & 5.0 \\
\bottomrule
\end{tabular*}
\end{table*}

\begin{table*}[!t]
\centering
\scriptsize
\setlength{\tabcolsep}{2pt}
\renewcommand{\arraystretch}{1.18}
\caption{NLL for the five complete FPT configurations: means over five folds, with sample SDs beneath. Row multipliers apply to means and SDs. Bold marks the lowest unrounded mean; underlined means are not significantly different from it (two-sided corrected resampled $t$-test over the five folds \citep{nadeau2003inference}, Holm-adjusted within the row \citep{holm1979simple}, $\alpha=0.05$). The final row averages ranks among these five configurations; its marks follow \Cref{tab:flow-ablation}. Smooth denotes automatic target smoothing.}
\label{tab:ablation-logloss}
\begin{tabular*}{\linewidth}{@{\extracolsep{\fill}}lrrrrr@{}}
\toprule
Dataset & Mixed + smooth & Exhaustive + smooth & Anchor + smooth & Mixed & Anchor \\
\midrule
Abalone & \shortstack[r]{\underline{1.971}\\[-1pt]{\fontsize{6}{7}\selectfont(0.051)}} & \shortstack[r]{\underline{1.970}\\[-1pt]{\fontsize{6}{7}\selectfont(0.060)}} & \shortstack[r]{\underline{2.020}\\[-1pt]{\fontsize{6}{7}\selectfont(0.058)}} & \shortstack[r]{\underline{0.923}\\[-1pt]{\fontsize{6}{7}\selectfont(0.402)}} & \shortstack[r]{\textbf{0.903}\\[-1pt]{\fontsize{6}{7}\selectfont(0.520)}} \\
Bike sharing & \shortstack[r]{\textbf{4.465}\\[-1pt]{\fontsize{6}{7}\selectfont(0.036)}} & \shortstack[r]{4.518\\[-1pt]{\fontsize{6}{7}\selectfont(0.054)}} & \shortstack[r]{4.793\\[-1pt]{\fontsize{6}{7}\selectfont(0.109)}} & \shortstack[r]{4.500\\[-1pt]{\fontsize{6}{7}\selectfont(0.035)}} & \shortstack[r]{4.965\\[-1pt]{\fontsize{6}{7}\selectfont(0.133)}} \\
California & \shortstack[r]{\underline{0.097}\\[-1pt]{\fontsize{6}{7}\selectfont(0.046)}} & \shortstack[r]{\underline{0.116}\\[-1pt]{\fontsize{6}{7}\selectfont(0.045)}} & \shortstack[r]{0.526\\[-1pt]{\fontsize{6}{7}\selectfont(0.098)}} & \shortstack[r]{\textbf{0.095}\\[-1pt]{\fontsize{6}{7}\selectfont(0.048)}} & \shortstack[r]{0.655\\[-1pt]{\fontsize{6}{7}\selectfont(0.217)}} \\
Communities & \shortstack[r]{\underline{-0.993}\\[-1pt]{\fontsize{6}{7}\selectfont(0.072)}} & \shortstack[r]{\textbf{-1.024}\\[-1pt]{\fontsize{6}{7}\selectfont(0.055)}} & \shortstack[r]{-0.896\\[-1pt]{\fontsize{6}{7}\selectfont(0.077)}} & \shortstack[r]{-0.932\\[-1pt]{\fontsize{6}{7}\selectfont(0.078)}} & \shortstack[r]{-0.793\\[-1pt]{\fontsize{6}{7}\selectfont(0.076)}} \\
Concrete & \shortstack[r]{\textbf{2.735}\\[-1pt]{\fontsize{6}{7}\selectfont(0.139)}} & \shortstack[r]{\underline{2.790}\\[-1pt]{\fontsize{6}{7}\selectfont(0.139)}} & \shortstack[r]{3.576\\[-1pt]{\fontsize{6}{7}\selectfont(0.213)}} & \shortstack[r]{\underline{2.742}\\[-1pt]{\fontsize{6}{7}\selectfont(0.136)}} & \shortstack[r]{3.550\\[-1pt]{\fontsize{6}{7}\selectfont(0.206)}} \\
Energy & \shortstack[r]{\textbf{0.348}\\[-1pt]{\fontsize{6}{7}\selectfont(0.090)}} & \shortstack[r]{\underline{0.393}\\[-1pt]{\fontsize{6}{7}\selectfont(0.104)}} & \shortstack[r]{\underline{0.645}\\[-1pt]{\fontsize{6}{7}\selectfont(0.213)}} & \shortstack[r]{\underline{0.406}\\[-1pt]{\fontsize{6}{7}\selectfont(0.084)}} & \shortstack[r]{\underline{1.401}\\[-1pt]{\fontsize{6}{7}\selectfont(0.438)}} \\
Kin8nm & \shortstack[r]{\underline{-0.661}\\[-1pt]{\fontsize{6}{7}\selectfont(0.024)}} & \shortstack[r]{-0.535\\[-1pt]{\fontsize{6}{7}\selectfont(0.031)}} & \shortstack[r]{0.122\\[-1pt]{\fontsize{6}{7}\selectfont(0.166)}} & \shortstack[r]{\textbf{-0.664}\\[-1pt]{\fontsize{6}{7}\selectfont(0.031)}} & \shortstack[r]{0.163\\[-1pt]{\fontsize{6}{7}\selectfont(0.231)}} \\
Naval & \shortstack[r]{\textbf{-5.782}\\[-1pt]{\fontsize{6}{7}\selectfont(0.015)}} & \shortstack[r]{-5.733\\[-1pt]{\fontsize{6}{7}\selectfont(0.034)}} & \shortstack[r]{-4.599\\[-1pt]{\fontsize{6}{7}\selectfont(0.157)}} & \shortstack[r]{-5.688\\[-1pt]{\fontsize{6}{7}\selectfont(0.026)}} & \shortstack[r]{-3.678\\[-1pt]{\fontsize{6}{7}\selectfont(0.087)}} \\
Protein & \shortstack[r]{\textbf{1.883}\\[-1pt]{\fontsize{6}{7}\selectfont(0.018)}} & \shortstack[r]{2.082\\[-1pt]{\fontsize{6}{7}\selectfont(0.025)}} & \shortstack[r]{2.527\\[-1pt]{\fontsize{6}{7}\selectfont(0.021)}} & \shortstack[r]{\underline{1.892}\\[-1pt]{\fontsize{6}{7}\selectfont(0.015)}} & \shortstack[r]{2.585\\[-1pt]{\fontsize{6}{7}\selectfont(0.075)}} \\
Power plant & \shortstack[r]{\textbf{2.528}\\[-1pt]{\fontsize{6}{7}\selectfont(0.020)}} & \shortstack[r]{2.685\\[-1pt]{\fontsize{6}{7}\selectfont(0.023)}} & \shortstack[r]{3.244\\[-1pt]{\fontsize{6}{7}\selectfont(0.050)}} & \shortstack[r]{\underline{2.530}\\[-1pt]{\fontsize{6}{7}\selectfont(0.023)}} & \shortstack[r]{3.354\\[-1pt]{\fontsize{6}{7}\selectfont(0.194)}} \\
\midrule Mean rank & \textbf{1.6} & 2.6 & 4.2 & \underline{2.1} & \underline{4.5} \\
\bottomrule
\end{tabular*}
\end{table*}

\begin{table*}[!t]
\centering
\scriptsize
\setlength{\tabcolsep}{2pt}
\renewcommand{\arraystretch}{1.18}
\caption{PCE for the five complete FPT configurations: means over five folds, with sample SDs beneath. Row multipliers apply to means and SDs. Bold marks the lowest unrounded mean; underlined means are not significantly different from it (two-sided corrected resampled $t$-test over the five folds \citep{nadeau2003inference}, Holm-adjusted within the row \citep{holm1979simple}, $\alpha=0.05$). The final row averages ranks among these five configurations; its marks follow \Cref{tab:flow-ablation}. Smooth denotes automatic target smoothing.}
\label{tab:ablation-pce1}
\begin{tabular*}{\linewidth}{@{\extracolsep{\fill}}lrrrrr@{}}
\toprule
Dataset & Mixed + smooth & Exhaustive + smooth & Anchor + smooth & Mixed & Anchor \\
\midrule
Abalone & \shortstack[r]{0.110\\[-1pt]{\fontsize{6}{7}\selectfont(0.021)}} & \shortstack[r]{0.109\\[-1pt]{\fontsize{6}{7}\selectfont(0.022)}} & \shortstack[r]{0.109\\[-1pt]{\fontsize{6}{7}\selectfont(0.022)}} & \shortstack[r]{\textbf{0.018}\\[-1pt]{\fontsize{6}{7}\selectfont(0.013)}} & \shortstack[r]{\underline{0.018}\\[-1pt]{\fontsize{6}{7}\selectfont(0.011)}} \\
Bike sharing & \shortstack[r]{\underline{0.018}\\[-1pt]{\fontsize{6}{7}\selectfont(0.006)}} & \shortstack[r]{\underline{0.019}\\[-1pt]{\fontsize{6}{7}\selectfont(0.006)}} & \shortstack[r]{0.027\\[-1pt]{\fontsize{6}{7}\selectfont(0.005)}} & \shortstack[r]{\textbf{0.008}\\[-1pt]{\fontsize{6}{7}\selectfont(0.003)}} & \shortstack[r]{\underline{0.012}\\[-1pt]{\fontsize{6}{7}\selectfont(0.003)}} \\
California & \shortstack[r]{\underline{0.016}\\[-1pt]{\fontsize{6}{7}\selectfont(0.005)}} & \shortstack[r]{\underline{0.013}\\[-1pt]{\fontsize{6}{7}\selectfont(0.003)}} & \shortstack[r]{\underline{0.024}\\[-1pt]{\fontsize{6}{7}\selectfont(0.007)}} & \shortstack[r]{\textbf{0.011}\\[-1pt]{\fontsize{6}{7}\selectfont(0.004)}} & \shortstack[r]{\underline{0.015}\\[-1pt]{\fontsize{6}{7}\selectfont(0.004)}} \\
Communities & \shortstack[r]{\underline{0.045}\\[-1pt]{\fontsize{6}{7}\selectfont(0.017)}} & \shortstack[r]{\underline{0.047}\\[-1pt]{\fontsize{6}{7}\selectfont(0.015)}} & \shortstack[r]{\underline{0.063}\\[-1pt]{\fontsize{6}{7}\selectfont(0.014)}} & \shortstack[r]{\textbf{0.028}\\[-1pt]{\fontsize{6}{7}\selectfont(0.011)}} & \shortstack[r]{\underline{0.036}\\[-1pt]{\fontsize{6}{7}\selectfont(0.008)}} \\
Concrete & \shortstack[r]{\underline{0.043}\\[-1pt]{\fontsize{6}{7}\selectfont(0.009)}} & \shortstack[r]{\underline{0.053}\\[-1pt]{\fontsize{6}{7}\selectfont(0.012)}} & \shortstack[r]{0.107\\[-1pt]{\fontsize{6}{7}\selectfont(0.015)}} & \shortstack[r]{\textbf{0.031}\\[-1pt]{\fontsize{6}{7}\selectfont(0.010)}} & \shortstack[r]{0.094\\[-1pt]{\fontsize{6}{7}\selectfont(0.020)}} \\
Energy & \shortstack[r]{\textbf{0.034}\\[-1pt]{\fontsize{6}{7}\selectfont(0.014)}} & \shortstack[r]{\underline{0.042}\\[-1pt]{\fontsize{6}{7}\selectfont(0.025)}} & \shortstack[r]{\underline{0.058}\\[-1pt]{\fontsize{6}{7}\selectfont(0.022)}} & \shortstack[r]{\underline{0.036}\\[-1pt]{\fontsize{6}{7}\selectfont(0.009)}} & \shortstack[r]{\underline{0.048}\\[-1pt]{\fontsize{6}{7}\selectfont(0.013)}} \\
Kin8nm & \shortstack[r]{\underline{0.037}\\[-1pt]{\fontsize{6}{7}\selectfont(0.007)}} & \shortstack[r]{\textbf{0.030}\\[-1pt]{\fontsize{6}{7}\selectfont(0.004)}} & \shortstack[r]{\underline{0.047}\\[-1pt]{\fontsize{6}{7}\selectfont(0.005)}} & \shortstack[r]{0.034\\[-1pt]{\fontsize{6}{7}\selectfont(0.004)}} & \shortstack[r]{\underline{0.039}\\[-1pt]{\fontsize{6}{7}\selectfont(0.004)}} \\
Naval & \shortstack[r]{0.216\\[-1pt]{\fontsize{6}{7}\selectfont(0.003)}} & \shortstack[r]{0.212\\[-1pt]{\fontsize{6}{7}\selectfont(0.007)}} & \shortstack[r]{0.206\\[-1pt]{\fontsize{6}{7}\selectfont(0.009)}} & \shortstack[r]{0.047\\[-1pt]{\fontsize{6}{7}\selectfont(0.002)}} & \shortstack[r]{\textbf{0.029}\\[-1pt]{\fontsize{6}{7}\selectfont(0.005)}} \\
Protein & \shortstack[r]{0.020\\[-1pt]{\fontsize{6}{7}\selectfont(0.002)}} & \shortstack[r]{\textbf{0.007}\\[-1pt]{\fontsize{6}{7}\selectfont(0.002)}} & \shortstack[r]{0.014\\[-1pt]{\fontsize{6}{7}\selectfont(0.002)}} & \shortstack[r]{0.016\\[-1pt]{\fontsize{6}{7}\selectfont(0.002)}} & \shortstack[r]{0.010\\[-1pt]{\fontsize{6}{7}\selectfont(0.001)}} \\
Power plant & \shortstack[r]{0.039\\[-1pt]{\fontsize{6}{7}\selectfont(0.007)}} & \shortstack[r]{\underline{0.025}\\[-1pt]{\fontsize{6}{7}\selectfont(0.006)}} & \shortstack[r]{0.036\\[-1pt]{\fontsize{6}{7}\selectfont(0.005)}} & \shortstack[r]{0.036\\[-1pt]{\fontsize{6}{7}\selectfont(0.005)}} & \shortstack[r]{\textbf{0.024}\\[-1pt]{\fontsize{6}{7}\selectfont(0.007)}} \\
\midrule Mean rank & 3.6 & \underline{2.8} & 4.3 & \textbf{1.8} & \underline{2.5} \\
\bottomrule
\end{tabular*}
\end{table*}

\begin{table*}[!t]
\centering
\small
\setlength{\tabcolsep}{2pt}
\renewcommand{\arraystretch}{1.18}
\caption{Mean seconds per outer fold over ten datasets. All configurations use fixed hyperparameters. Fit, prediction and metric evaluation are separate recorded phases; averages describe the full evaluation pipeline.}
\label{tab:ablation-runtime}
\begin{tabular*}{\linewidth}{@{\extracolsep{\fill}}lrrr@{}}
\toprule
Configuration & Fit (s) & Predict (s) & Metrics (s) \\
\midrule
Mixed + smooth & 40.7 & 11.6 & 0.9 \\
Exhaustive + smooth & 49.6 & 19.7 & 1.0 \\
Anchor + smooth & 35.2 & 2.8 & 0.5 \\
Mixed & 20.6 & 61.1 & 353.7 \\
Anchor & 19.0 & 10.5 & 14.4 \\
\bottomrule
\end{tabular*}
\end{table*}

\FloatBarrier
\subsection{Sensitivity to the conditional distribution}
\label{app:results-synthetic}

\paragraph{Conditional mean and noise.}
We draw independent uniform covariates on $[-1,1]$ and generate each
response as
\begin{equation}
\label{eq:synthetic-response}
Y=\sqrt{r^2}\,g(X)+\sqrt{1-r^2}\,s_H(X_1)\,\varepsilon.
\end{equation}
The first $d$ covariates determine the signal and the remaining
$p_{\rm irr}$ covariates are independent of the response. For even $d$, we use
\[
g(x)=\frac{\sqrt{2}\sum_{j=1}^{d}\sin(\pi x_j)
             +3\sum_{j=1}^{d/2}x_{2j-1}x_{2j}}{\sqrt{3d/2}}.
\]
Under independent uniform covariates, the weighted main effects and
interactions contribute variances $d$ and $d/2$ to the numerator and
have zero covariance. The signal therefore has mean zero and variance one
for every $d$ in the experiment. Only $X_1$ controls the noise scale.
To vary the conditional standard
deviation while keeping the average noise variance fixed, we set
\[
s_H(x_1)=\frac{H^{x_1/2}}{\sqrt{C_H}},\qquad
C_H=\EE[H^{X_1}]=
\begin{cases}
1, & H=1,\\
(H-H^{-1})/(2\log H), & H>1.
\end{cases}
\]
Thus $\EE[s_H(X_1)^2]=1$. At $H=1$ the scale is constant, and at $H>1$
the conditional standard deviation increases by a factor of $H$ from
$X_1=-1$ to $X_1=1$. For finite $\nu$ we draw $T_\nu$ from the
standard Student-$t$ distribution and set
$\varepsilon=\sqrt{(\nu-2)/\nu}\,T_\nu$. For $\nu=\infty$ we draw
$\varepsilon$ from the standard normal distribution.
The error $\varepsilon$ is independent of the covariates and has unit
variance. These normalizations give the response unit variance, with
$r^2$ the fraction explained by the conditional mean. The conditional
CDF is known at every test point.

\paragraph{Settings and replication.}
Each sweep in \cref{fig:syn-sweeps} varies a single factor over the
values below, with the reference value in bold.
\begin{center}
\begin{tabular}{@{}ll@{}}
\toprule
Factor & Values \\
\midrule
Explained variance $r^2$ & $0.1,\ 0.3,\ \mathbf{0.6},\ 0.9$ \\
Conditional standard deviation ratio $H$ & $1,\ \mathbf{2},\ 4,\ 8$ \\
Noise degrees of freedom $\nu$ & $3,\ 5,\ \mathbf{10},\ \infty$ \\
Relevant covariates $d$ & $2,\ \mathbf{4},\ 8,\ 16$ \\
Irrelevant covariates $p_{\rm irr}$ & $0,\ \mathbf{10},\ 50,\ 100$ \\
Training observations $n$ & $500,\ 1{,}000,\ \mathbf{2{,}000},\ 5{,}000,\ 10{,}000$ \\
\bottomrule
\end{tabular}
\end{center}
Counting the shared reference once gives 20 settings. Eight additional
settings combine $r^2\in\{0.1,0.9\}$, $H\in\{1,8\}$ and
$\nu\in\{3,\infty\}$, keeping $d$, $p_{\rm irr}$ and $n$ at their
reference values. For replicate $k=0,\ldots,4$, we draw $n$ training
observations with seed $2k$ and 2,000 test observations with seed
$2k+1$. Each covariate and the error have separate random streams.
Within a replicate, increasing $n$ preserves the earlier training rows,
and adding irrelevant covariates preserves the responses. In both the tuned
and fixed-default comparisons, the estimator seed is $10000+k$ and
Treeffuser's sampling seed is $20000+k$.

\paragraph{Tuned and fixed-default comparisons.}
For each setting and replicate, we tune FPT, Treeffuser and CatBoost-R
using the protocol and search spaces in \cref{app:hyperparameters}.
Tuned FPT monitors validation CRPS. In the fixed-default comparison,
FPT uses the FPT-D settings in \cref{tab:fpt-hyperparameters} with the shared
controls in \cref{app:fpt-configurations} and monitors validation NLL.
Default Treeffuser uses at most 3,000 boosting iterations,
30 noisy repeats, learning rate 0.1, 31 leaves, validation fraction
0.1 and patience 50; its predictive sampling is the same 256-draw,
50-step configuration used in the tuned synthetic comparison. Default
CatBoost uses 1,000 iterations with no early stopping. Its recorded
resolved settings are learning rate 0.03, depth 6, $\ell_2$ leaf
regularization 3 and minimum-variance sampling (MVS) bootstrapping; its objective remains
\texttt{RMSEWithUncertainty}. These fixed-default runs use one worker
per method.

\begin{figure}[!htbp]
\centering
\includegraphics[width=0.55\linewidth]{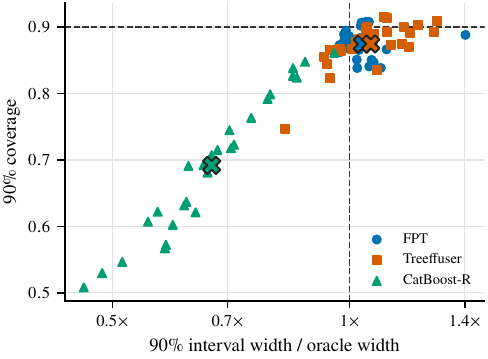}
\caption{Coverage and width of nominal 90\% prediction intervals for the three CRPS-tuned models. CatBoost-R uses RMSE early stopping. Each point summarizes five replicates in one setting. The cross marks arithmetic-mean coverage and geometric-mean relative width across settings. Dashed lines mark nominal coverage and the oracle width.}
\label{fig:syn-calibration}
\end{figure}

\begin{figure}[!htbp]
\centering
\includegraphics[width=\textwidth]{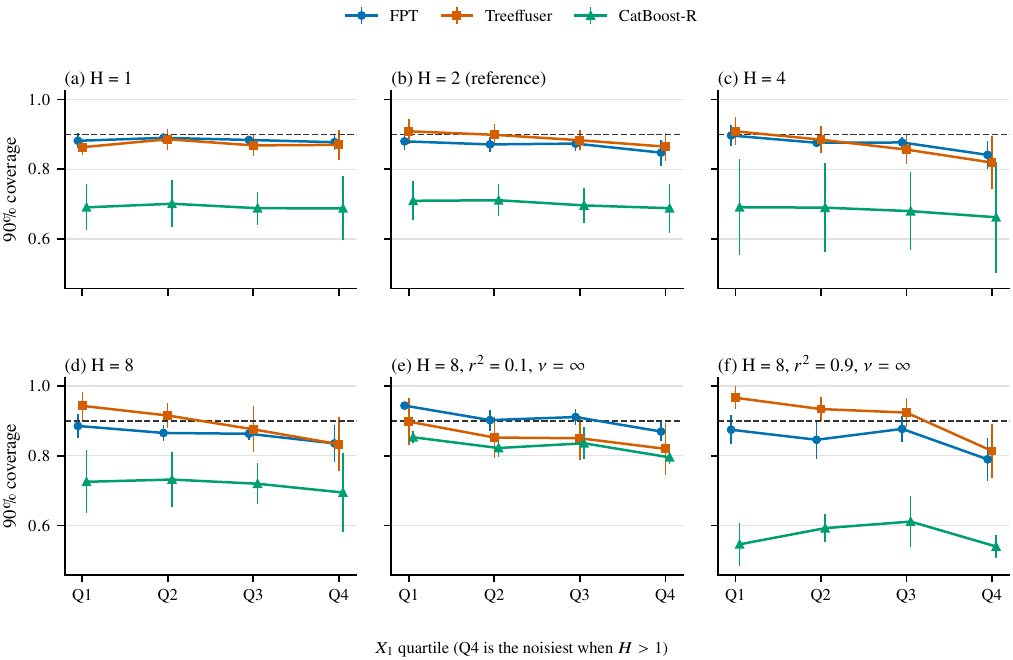}
\caption{90\% coverage by $X_1$ quartile along the heteroscedasticity sweep and at two corners with $H=8$, mean $\pm 1$ SD over five replicates. Q4 is the noisiest quartile when $H>1$. All three models are CRPS-tuned; CatBoost-R uses RMSE early stopping.}
\label{fig:syn-quartiles}
\end{figure}

\paragraph{Prediction interval coverage and width.}
\Cref{fig:syn-calibration} compares interval coverage and width across the
28 synthetic settings.
\Cref{fig:syn-quartiles} breaks down 90\% interval coverage by quartile
of $X_1$, the covariate that controls the noise scale.

\end{document}